\documentclass[runningheads]{llncs}
\pdfoutput=1
\usepackage{graphicx}
\usepackage{amsmath,amssymb} % define this before the line numbering.
\usepackage{color}
\usepackage[width=122mm,left=12mm,paperwidth=146mm,height=193mm,top=12mm,paperheight=217mm]{geometry}

\usepackage{times}
\usepackage{afterpage}
\usepackage{epsfig}
\usepackage{mathtools,empheq}
\usepackage{caption}
\usepackage{verbatim}
\usepackage{xspace}
\usepackage{pbox}
\usepackage{afterpage,placeins}

\usepackage{subfig}
\usepackage{multirow}
\usepackage{array}

\def\ReduceBeforeCaptionfigspace{}
\def\ReduceAfterCaptionfigspace{}

\newcommand{\avg}{\operatornamewithlimits{avg}}

\newcommand{\mccn}{\textsc{mccn}\xspace}

\newcommand{\Gama}{\boldsymbol{\Gamma}}
\newcommand{\gama}{\boldsymbol{\gamma}}

\newcommand{\ie}{\emph{i.e.}}
\newcommand{\eg}{\emph{e.g.}}

\begin{document}
% \renewcommand\thelinenumber{\color[rgb]{0.2,0.5,0.8}\normalfont\sffamily\scriptsize\arabic{linenumber}\color[rgb]{0,0,0}}
% \renewcommand\makeLineNumber {\hss\thelinenumber\ \hspace{6mm} \rlap{\hskip\textwidth\ \hspace{6.5mm}\thelinenumber}}
% \linenumbers
\pagestyle{headings}
\mainmatter
\def\ECCV16SubNumber{707}  % Insert your submission number here

\title{From Multiview Image Curves to 3D Drawings$\,^*$} % Replace with your title

\titlerunning{From Multiview Image Curves to 3D Drawings: Expanded Version}

\authorrunning{A.\ Usumezbas, R.\ Fabbri and B.\ B.\ Kimia}

\author{%
Anil~Usumezbas\\
SRI International\\
{\small\tt{anil.usumezbas@sri.com}}
\and\ \\[0.5em]
Ricardo~Fabbri\\
%State University of Rio de Janeiro\\
Polytechnic Institute\\
{\tt\small rfabbri@iprj.uerj.br}
% For a paper whose authors are all at the same institution,
% omit the following lines up until the closing brace
% Additional authors and addresses can be added with ``\and'',
% just like the second author.
% To save space, use either the email address or home page, not both
\and\ \\[0.5em]
Benjamin~B.~Kimia\\
%Brown University\\
Shool of Engineering\\
{\tt\small benjamin\_kimia@brown.edu}
}

\institute{SRI International, State University of Rio de Janeiro and Brown
  University\footnotetext{$\ ^*$ ECCV 2016, expanded version with tweaked figures and
    including an overview of the supplementary material available at \url{multiview-3d-drawing.sourceforge.net}.}}

\maketitle

\begin{abstract}
Reconstructing 3D scenes from multiple views has made impressive strides
in recent years, chiefly by correlating isolated feature points, intensity
patterns, or curvilinear structures. In the general setting --
without controlled acquisition, abundant texture, curves and
surfaces following specific models or limiting
scene complexity -- most methods produce unorganized point clouds,
meshes, or voxel representations, with some exceptions producing
unorganized clouds of 3D curve fragments. Ideally, many applications
require structured representations of curves, surfaces and their spatial
relationships. This paper presents a step in this direction by formulating an approach that combines 2D image curves into a collection of 3D curves, with topological connectivity between them represented as a 3D graph. This results in a \textbf{3D drawing}, which is complementary to surface representations in the same sense as a 3D scaffold complements a tent taut over it. We evaluate our results against truth on synthetic and real datasets.
\end{abstract}

%\vspace{-1cm}

\section{Introduction}

The automated 3D reconstruction of \emph{general} scenes from multiple views
obtained using conventional cameras, under uncontrolled 
acquisition, is a paramount goal of computer vision,
ambitious even by modern standards.
%Challenges relate to the large-scale
%choices of appropriate representations and techniques to deal simultaneously
%with wildly different materials (\eg, non-Lambertian), geometric models (\eg,
%general curved manifolds, discontinuities, textures, deformations, at different
%scales), region types (\eg, textured and textureless regions), unknown
%illumination conditions, shadows and shades, large viewpoint differences,
%background clutter, arbitrary number of objects and unknown camera parameters,
%to name a few.  
While a fully complete working system addressing all the underlying challenges
is beyond current technology, significant progress has been made in the past few
years using approaches that fall into three broad classes,
depending on whether one focuses on correlating isolated
points, surface patches, or curvilinear structures across views, as described
below.

\begin{figure}
  \hspace{-0.05cm}\begin{minipage}[c]{0.6\linewidth}
    \includegraphics[width=0.3\linewidth]{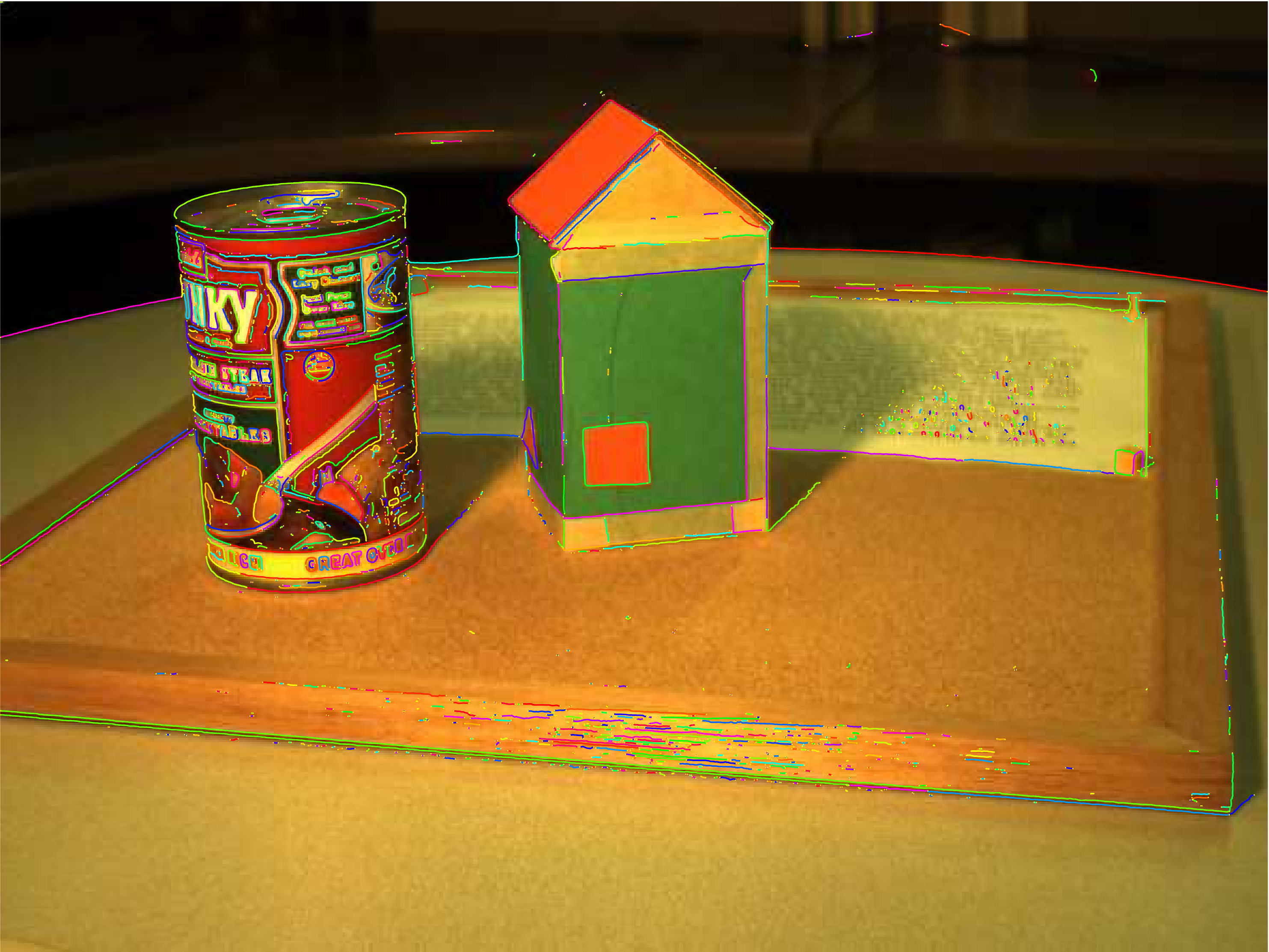}
    \includegraphics[width=0.3\linewidth]{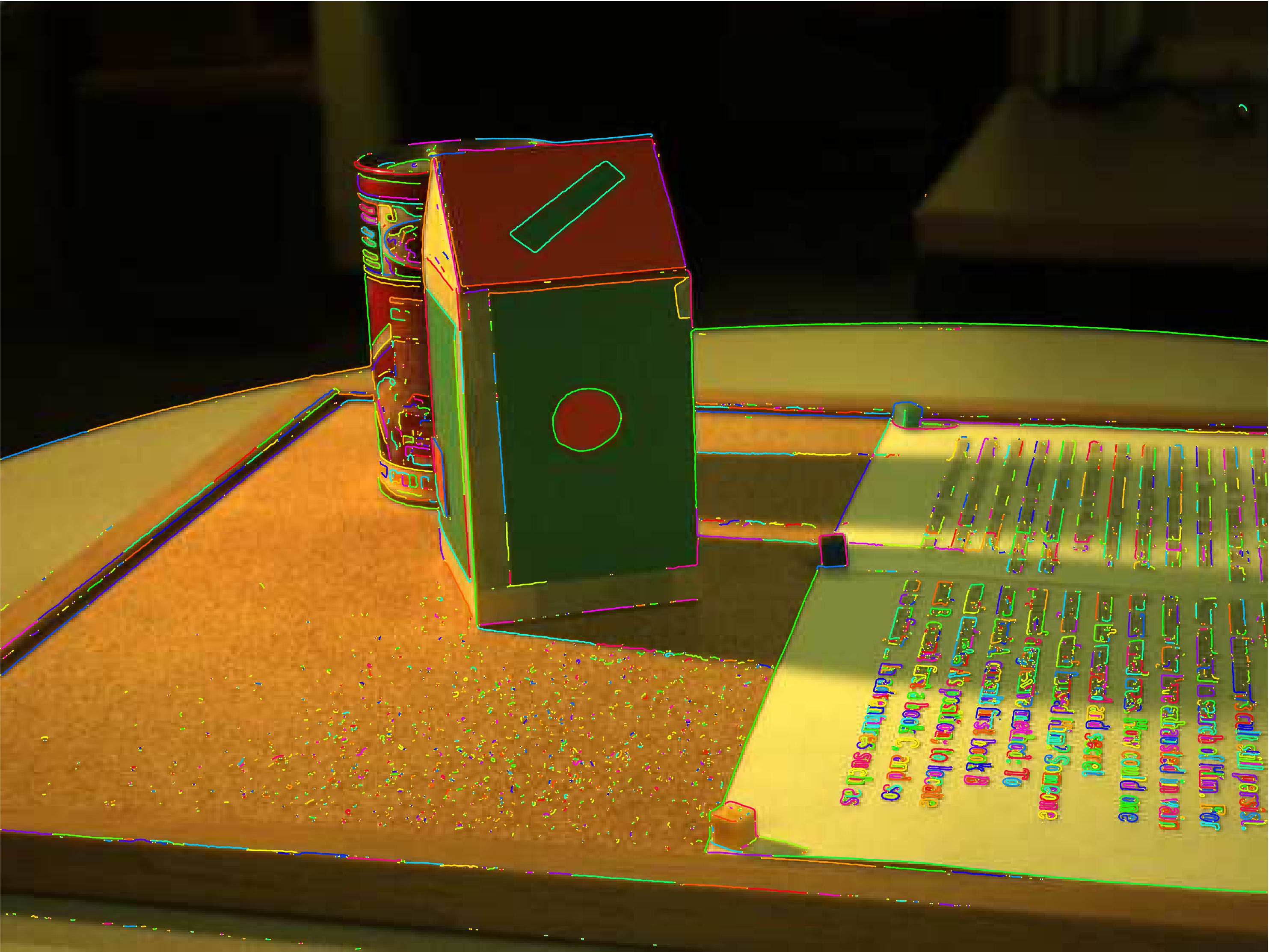}
    \includegraphics[width=0.3\linewidth]{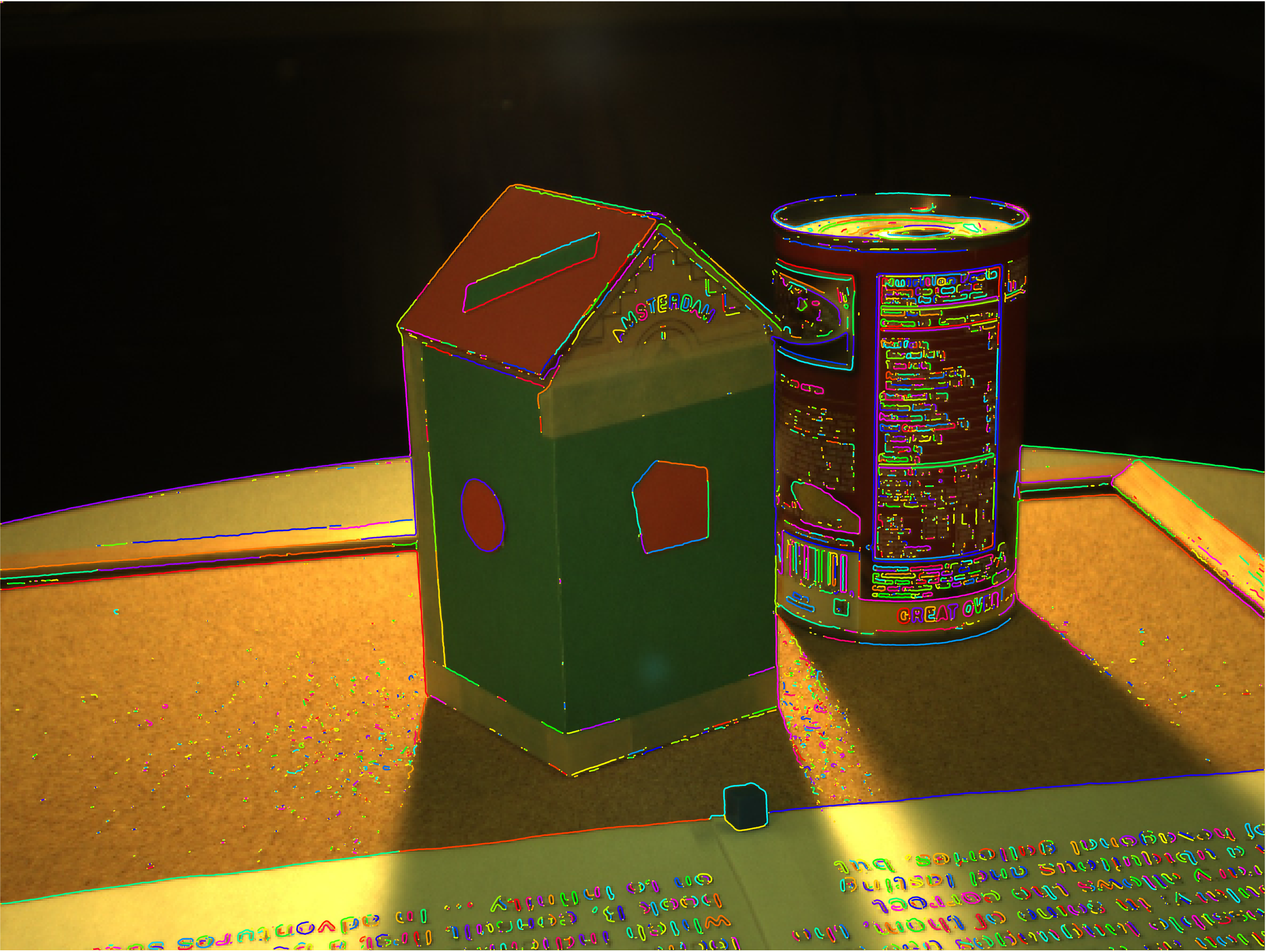}\\[0.5mm]
    \includegraphics[width=0.3\linewidth]{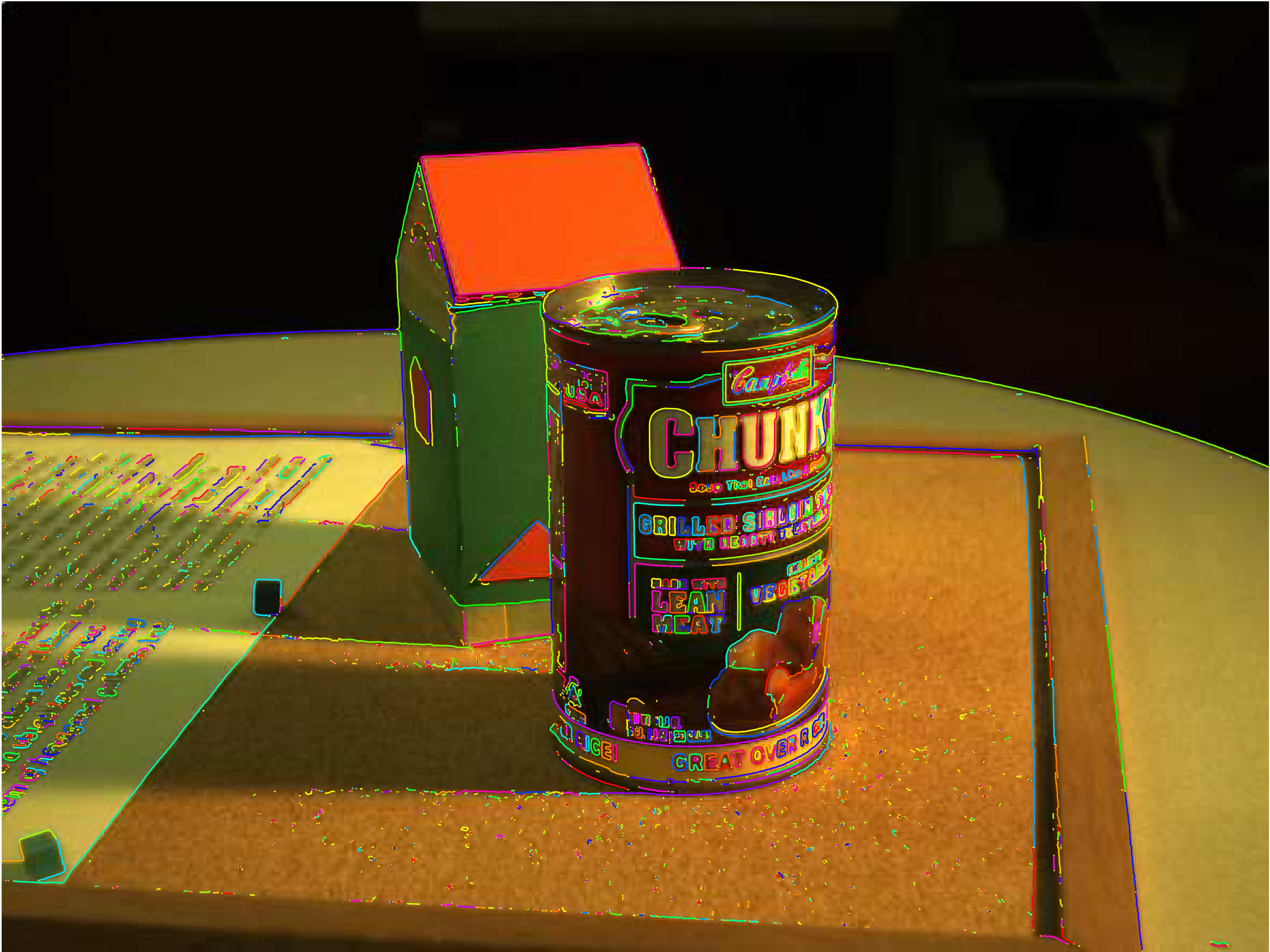}
    \includegraphics[width=0.3\linewidth]{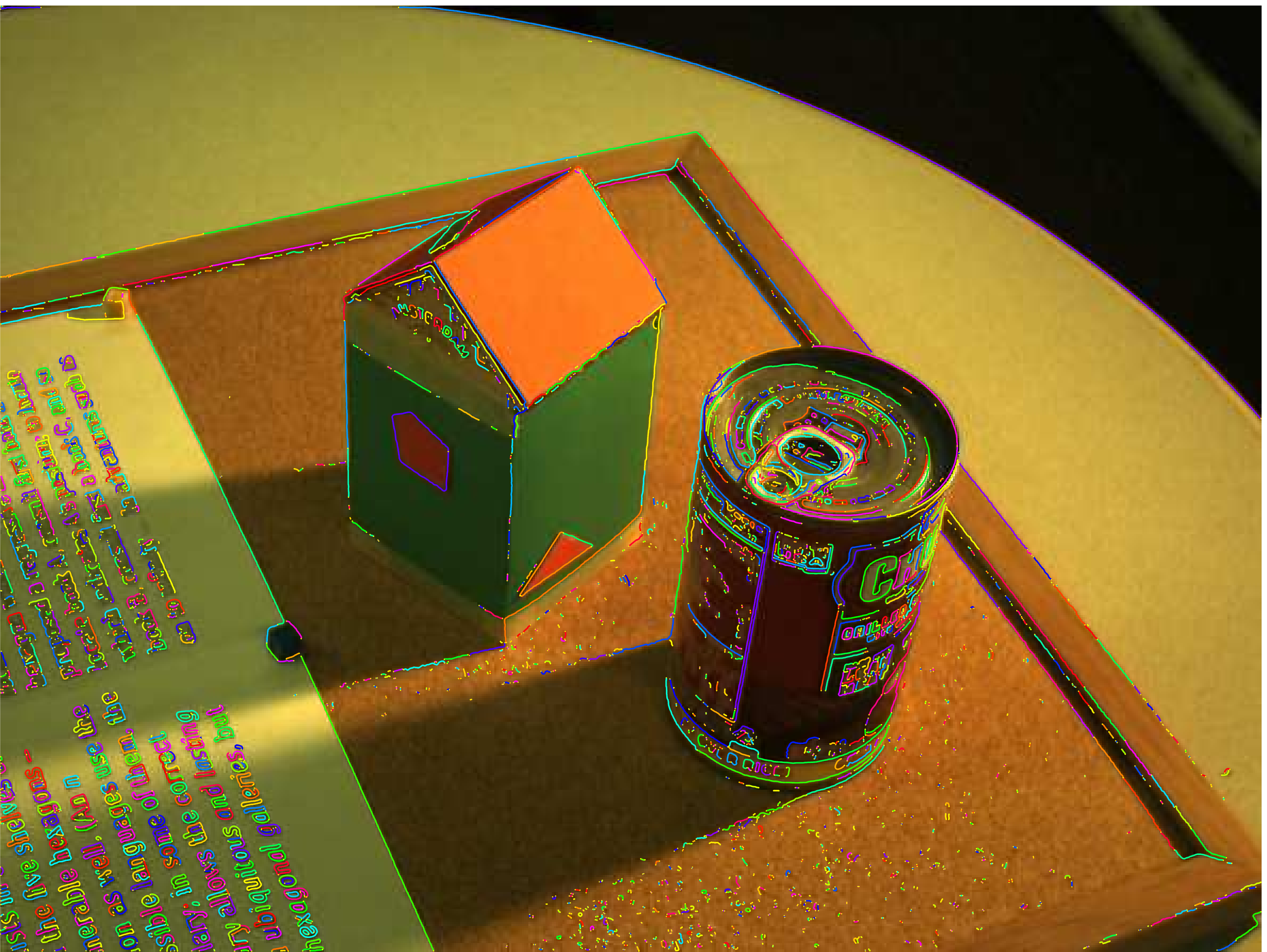}
    \includegraphics[width=0.3\linewidth]{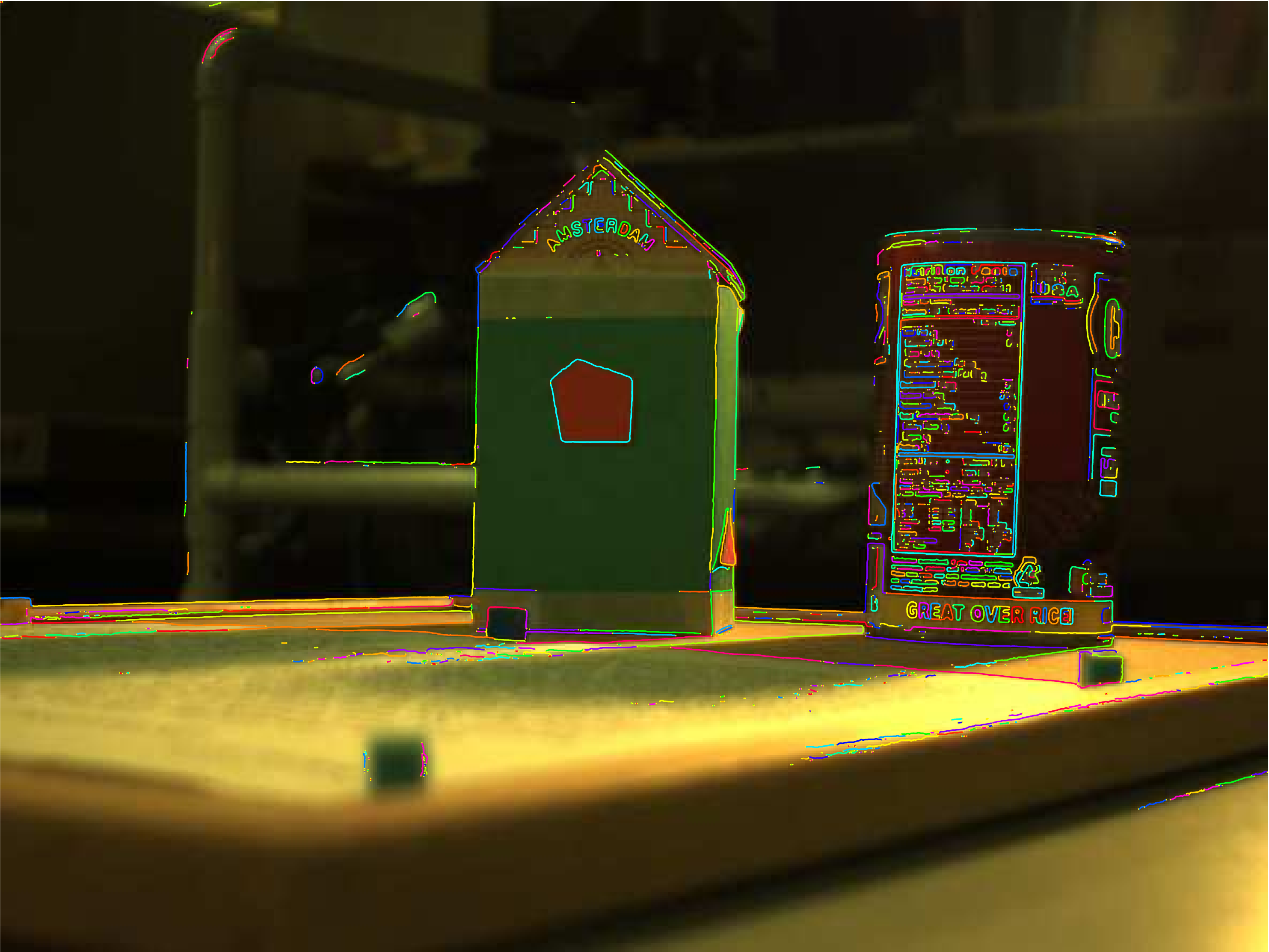}
  \end{minipage}
   \begin{minipage}[c]{0.4\linewidth}
     \vspace{-0.7cm}\hspace{-0.4cm}\includegraphics[width=1.1\linewidth]{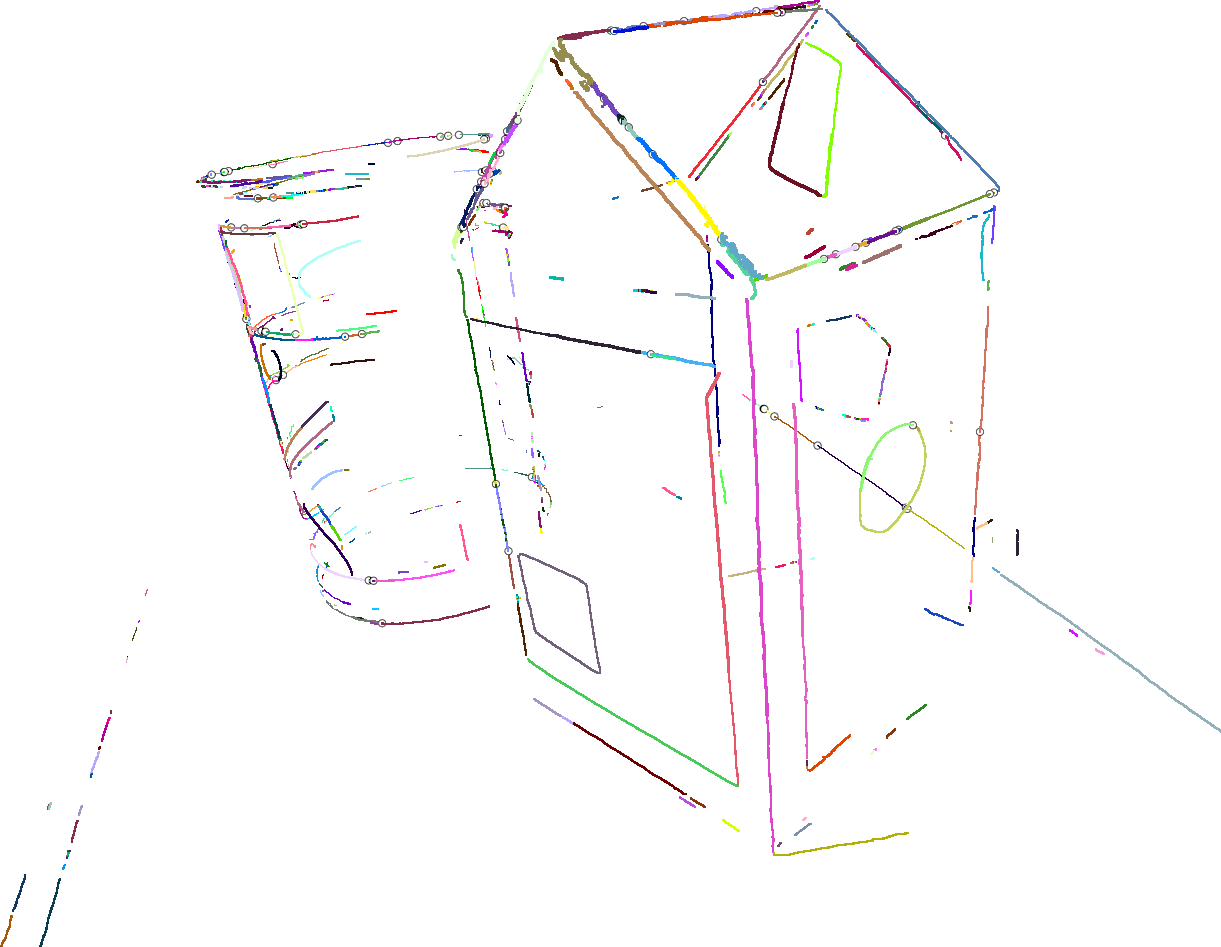}
  \end{minipage}
%  \vspace{1em}
	%\ReduceBeforeCaptionfigspace
	\caption{
		Our approach transforms calibrated views of a scene into a ``3D drawing'' --
    a graph of 3D curves meeting at junctions. Each
		curve is shown in a different color. 
    \emph{\small(Please zoom in to examine closely. The 3D model is available
    as supplementary data.)}}
	\label{fig:recon:results}
\end{figure}

A vast majority of multiview reconstruction methods rely on correlating isolated interest points across views to produce
an unorganized 3D cloud of points. 
The
\textbf{interest-point-based approach} has been highly successful in
reconstructing large-scale scenes with {\em texture-rich images}, in systems
such as in Phototourism and recent large-scale 3D reconstuction
work~\cite{Heinly:Frahm:etal:CVPR2015,Pollefeys:VanGool:etal:handheld:IJCV2004,Argarwal:Snavely:etal:ICCV09,Diskin:Vijayan:JEI2015}.
Despite their manifest usefulness, these methods generally cannot represent smooth, textureless regions (due to the sparsity of interest points in image regions with homogeneous appearance), or regions that change appearance 
drastically across views. 
%%% xxx put somewhere else, its out of place but important:
%even though contours clearly match. \indraftnote{Ben: we shouldn't change the previous
%sentence since it was to attend a reviewer criticism on our ECCV paper}
This limits their applicability, especially in
man-made environments~\cite{Simoes:etal:SVR2014} and objects such as
cars~\cite{Shinozuka:Saito:VRIC14}, non-Lambertian surfaces such as that of the
sea, appearance variation due to changing
weather~\cite{Baatz:Pollefeys:etal:ECCV12}, and wide
baseline~\cite{Moreels:Perona:IJCV07}.

Another approach matches
intensity patterns across views using multiview stereo, producing denser
point clouds or mesh reconstructions.
\textbf{Dense multi-view stereo} produces
detailed 3D reconstructions of objects imaged under controlled conditions by a
large number of precisely calibrated
cameras~\cite{Furukawa:Ponce:CVPR2007,Habbecke:Kobbelt:CVPR2007,Hernandez:Schmitt:CVIU04,Goesele:etal:ICCV07,Seitz:etal:CVPR06,Calakli:etal:3DIMPVT2012,Restrepo:etal:JPRS2014}.
For general, complex scenes with various kinds of objects and surface
properties, this approach has shown most promise towards obtaining an accurate
and dense 3D model of a given scene.  Homogeneous areas, such as walls of a
corridor, repeated texture, and areas with view-dependent intensities create
challenges for these methods.

%Another class of methods rely on optimizing either voxel occupancy [\textcolor{red}{reference needed to back up}] or a 
%global inverse-rendering function that optimizes for lighting and a single, dense surface simultaneously, given an initial 
%guess [\textcolor{red}{reference needed to back up}]. These approaches often need to be initialized by the visual hull of the object, a good initial guess or a bounded 3D voxel volume, all of which can create practical limitations for general scenery; (b) they typically make strong assumptions about the number of objects in the scene or their attributes (\eg man-made structures, planar surfaces, single object of uniform material etc.)
%(c) they tend to smooth out semantic details of the 
%scene, like ridges and corners; (d) they sometimes require controlled acquisition, or require camera calibration on an intensity level.

A smaller number of techniques correlate
and reconstruct image \textbf{curvilinear structure} across views, resulting in 3D curvilinear 
structure. Pipelines based on straight
lines (see~\cite{Lebeda:etal:ACCV2014,Zhang:line:PHDThesis2013,Fathi:etal:AEI2015} for recent reviews),
algebraic and general curve features~\cite{Teney:Piater:3DIMPVT12,Litvinov:etal:IC3D2012,Fabbri:Kimia:CVPR10,Fabbri:Giblin:Kimia:ECCV12,Wendel:etal:CVWW2011,Berthilsson:etal:IJCV2001,Fabbri:Kimia:EMMCVPR2005}
have been proposed, but some lack generality, \eg, requiring 
specific curve models~\cite{Carrasco:etal:LNCS2012}.
The 3D Curve Sketch system~\cite{Fabbri:PhD:2010,Fabbri:Giblin:Kimia:ECCV12,Fabbri:Kimia:CVPR10} operates on
multiple views by pairing curves from two arbitrary ``hypothesis views'' at a
time via epipolar-geometric consistency. A curve pair 
reconstructs to a 3D curve fragment hypothesis, whose reprojection onto several
other ``confirmation views'' gathers support from subpixel 2D edges.
The curve pair hypotheses with enough support
result in an unorganized set of 3D curve fragments, the ``3D Curve Sketch''. 
While the resulting 3D curve segments are visually appealing, they are
fragmented, redundant, and lack explicit inter-curve organization.

\begin{figure}[t]
	\centering
	\includegraphics[width=1\linewidth]{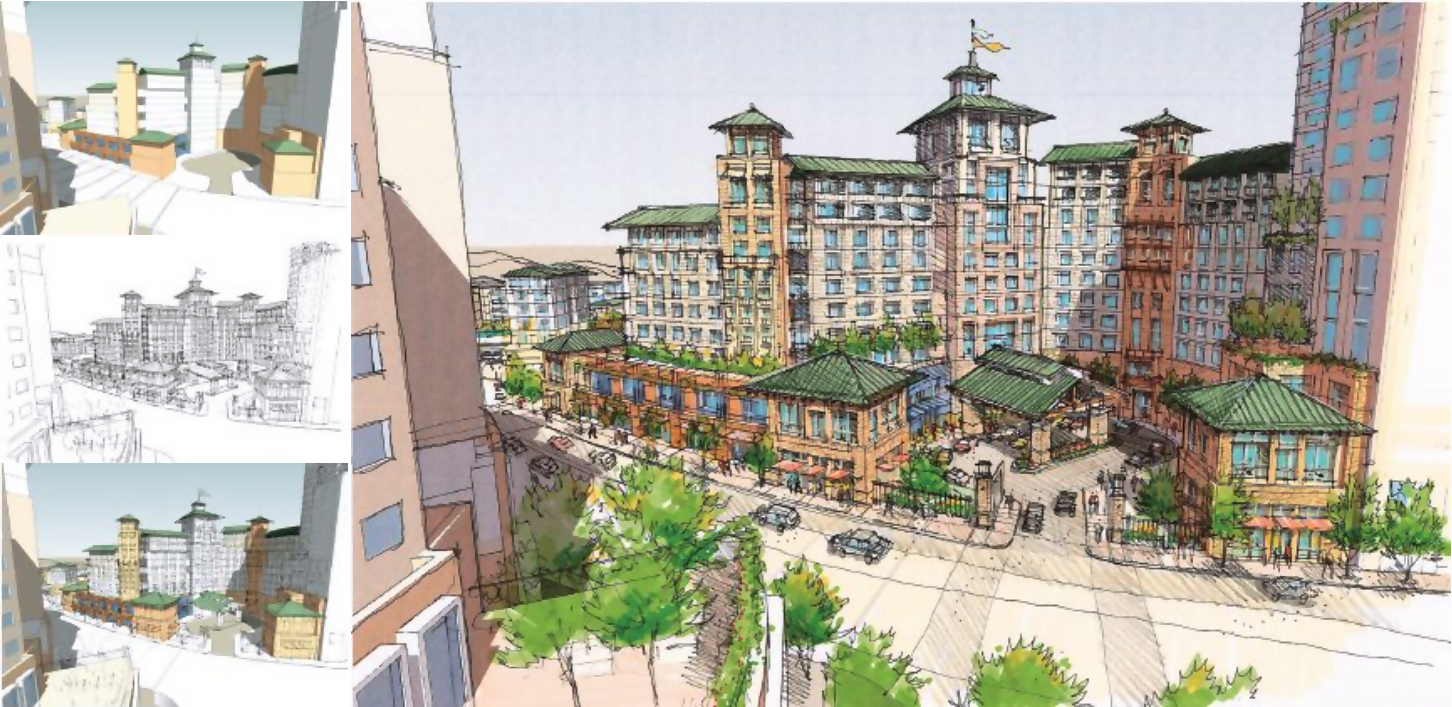}
	\ReduceBeforeCaptionfigspace
	\caption{%
		\small
		\textbf{3D drawings for urban planning and industrial design.} A
		process from professional practice for communicating
		solution concepts with a blend of computer and
		handcrafted renderings~\cite{Leggitt:drawing:book,Yee:architectural:book}.
    New designs are often based off real object references,
    mockups or massing models for selecting viewpoints and rough shapes.
    These can be modeled \emph{manually} in, \eg, Google
    Sketchup (top-left), in some cases from reference imagery. The desired 2D views are rendered
    and \emph{manually} traced into a reference curve sketch (center-left,
    bottom-left) easily modifiable
    to the designer's vision. The stylized drawings to be
    presented to a client are often produced by \emph{manually} tracing
    and painting over the reference sketch (right). Our system can be used to generate
    reference \emph{3D curve drawings} from video footage of the real site for
    urban planning, saving manual
		interaction, providing initial information such as
    rough dimensions, and aiding the selection of pose, editing and
    tracing. The condensed 3D curve drawings make room for the
    artist to overlay his concept and harness imagery as a clean reference,
    clear from details to be redesigned.
	}\label{fig:architectural:drawing}
	\ReduceAfterCaptionfigspace
\end{figure}

The plethora of multiview representations, as documented above, arise because 3D
structures are geometrically and semantically
rich~\cite{Zia:Stark:Schindler:IJCV2015,Feng:Medioni:etal:SIGGRAPH14}.  A
building, for example, has walls, windows, doorways, roof, chimneys, etc. The
structure can be represented by sample points (\ie, unorganized
cloud of points) or a surface mesh where connectivity among points is captured.
This representation, especially when rendered with surface albedo or texture, is
visually appealing. However, the representation also leaves out a great deal of
semantic information: which points or mesh areas represent a window or a wall?
Which two walls are adjacent? The representation of such components, or parts,
requires an explicit representation of part boundaries such as ridges, as well
as where these boundaries come together, such as junctions.

The same point can equally arise if objects in the scene were solely defined
by their curve structures. A representation of a building by its ridges may
usually give an appealing impression of its structure, but it fails to identify
the walls, \ie, which collection of 3D curves bound a wall and what its geometry
is. Both surfaces and curves are important and needed across the board, \eg, in
applications such as robotics~\cite{Carlson:etal:ICRA2014}, urban planning and
industrial design~\cite{Yee:architectural:book,Leggitt:drawing:book}, Fig.~\ref{fig:architectural:drawing}.
%It is in this spirit that the current
%work on extraction and representation of 3D curve structure is presented. 

In general, image curve fragments are attractive because they have
good localization, they have greater invariance than interest points to changes
in illumination, are stable over a greater range of baselines, and are typically
denser than interest points.  Furthermore, the reflectance or ridge curves
provide boundary condition for surface reconstruction, while occluding contour
variations across views lead to surfaces~\cite{Giblin:Motion:Book,Liu:Cooper:etal:PAMI07,Crispell:etal:LNCS2009}.
Recent studies strongly support the notion that image curves contain much of the
image
information~\cite{Koenderink:Wagemans:etal:iPerception2013,Zucker:PIEEE2014,Kunsberg:Zucker:LNM2014,Cole:etal:SIGGRAPH09}.
Moreover, curves are structurally rich as reflected by their differential
geometry, a fact which is exploited both in recent computer
systems~\cite{Zucker:PIEEE2014,Abuhashim:Sukkarieh:IROS2012,Fabbri:Giblin:Kimia:ECCV12,Fabbri:Kimia:CVPR10}
and peception studies~\cite{Fleming:etal:PNAS2011,Zucker:PIEEE2014}.

This paper develops the technology to process a series of (intrinsic and
extrinsically) calibrated multiview images to generate a \emph{3D curve drawing}
as a graph of 3D curve segments meeting at junctions. The ultimate goal of this
approach is to integrate the 3D curve drawing with the traditional recovery of
surfaces so that 3D curves bound the 3D curve segments, towards a more semantic
representation of 3D structures. The 3D curve drawing can also be of
independent value in applications such as fast recognition of general 3D
scenery~\cite{Wendel:etal:CVWW2011}, efficient transmission of general 3D
scenes, scene understanding and modeling by reasoning at
junctions~\cite{Mattingly:etal:JVLC2015}, consistent non-photorealistic
rendering from video~\cite{Chen:Klette:IVT2014}, modeling of branching
structures, among others~\cite{Rao:etal:IROS2012,Kowdle:etal:ECCV10,Ruizhe:Medioni:CVPR2014}.

The paper is organized as follows. In Section~\ref{sec:improved:sketch} we
review the 3D curve sketch, identify three shortcomings and
suggest solutions to each, resulting in the \emph{Enhanced Curve Sketch}. Since
the original 3D curve sketch was built around a few views at a time, it did not
address fundamental issues surrounding integration of information from numerous
views. Section~\ref{sec:3d:drawing} presents as our main contribution
the multiview integration of information both at edge- and curve-level,
which naturally leads to junctions.
Section~\ref{sec:results} validates the approach using real and synthetic
datasets.

%The main contribution of the proposed approach is the unprecedented structural
%richness of the proposed 3D reconstruction that is applicable to general %scenes,
%enabled by leveraging multiple view consistency of image curve fragments and
%junctions.  The proposed approach begins by tackling three important
%shortcomings of the 3D
%curve sketch representation, creating what we refer to as an \emph{Enhanced 
%3D Curve Sketch}, Section~\ref{sec:improved:sketch}. This in turn, serves as 
%the basis for forming a 3D curve drawing where long and meaningful curves come %together, with their spatial relationship captured in a graph, %Section~\ref{sec:3d:drawing}. The 3D curve
%drawing is evaluated on several datasets, showing impressive reconstruction %quality, Section~\ref{sec:results}.

\nocite{
	Kowdle:etal:ECCV10,
	Chen:Klette:IVT2014,
	Zhu:Luong:etal:CVPR2014,
	Rao:etal:IROS2012,
	Abuhashim:Sukkarieh:IROS2012,
	Yang:Ahuja:TIP2012,
	Chen:Klette:IVT2014,
	Simoes:etal:SVR2014,
	Kanazawa:Kanatani:etal:IPSJTCVA2014,
	Heinly:Frahm:etal:CVPR2015,
	Teney:Piater:3DIMPVT12,
	Zhang:line:PHDThesis2013,
	Lebeda:etal:ACCV2014,
	Bertasius:Shi:Torresani:CVPR2015,
	Fleming:etal:PNAS2011,
	Kunsberg:Zucker:LNM2014,
	Zucker:PIEEE2014,
	Koenderink:Wagemans:etal:iPerception2013,
	Ruizhe:Medioni:CVPR2014,
	%Carrasco:etal:LNCS2012,
	Shinozuka:Saito:VRIC14,
	Fathi:etal:AEI2015,
	%Zia:Stark:Schindler:IJCV2015,
	%Feng:Medioni:etal:SIGGRAPH14,
	Restrepo:etal:JPRS2014,
	Calakli:etal:3DIMPVT2012,
	% eval:
	Strecha:etal:CVPR2008,
	Jensen:etal:CVPR14}

%\vspace{-0.5cm}
\section{Enhanced 3D Curve Sketch}
\label{sec:improved:sketch}

%\vspace{-1.2cm}
%  \begin{table}[h]
%  	\renewcommand{\arraystretch}{1.25}
%  	\caption{Table of notation.}
%  %  \vspace{-0.2cm}
%  	\begin{center}
%  		\footnotesize
%  		\begin{tabular}{| c | p{9cm} |}
%  			\hline
%  			\multicolumn{1}{|c|}{\textbf{Symbol}} &
%  			\multicolumn{1}{c|}{\textbf{Description}} \\\hline\hline
%  			$\mathcal V = V^1,\dots,V^N$ & A set of images, views of a common scene\\ \hline %    $P^i$ & Projection matrix calibrating views $V_i$ to the world coordinate system  \\ \hline
%  			%    $V^v$ & \pbox{5cm}{A set of images, views of a common scene,\\
%  			%    $v=1,...,N$.} \\ \hline %    $P^i$ & Projection matrix calibrating views $V_i$ to the world coordinate system  \\ \hline
%  			%    $e_i^m$ & Edge $m$ in image $i$ \\ \hline
%  			$\mathcal E^v$ & Edge map of the image $V^v$ \\ \hline
%  			$\gama_l^v$ & Curve fragment data at view $v$, $l=1,...,M^v$\\ \hline
%  			$\Gama_k$ & Reconstructed 3D curve, $k=1,\dots,K$ \\ \hline
%  			$\gama^{k,v}$ & The reprojection of $\Gama_k$ onto each view $v$ \\\hline
%  			$\phi_{i, j}(s_i,s_j)$ & A weighted corresp. link between 3D curve points
%  			$\Gama_i(s_i)$ and $\Gama_j(s_j)$\\ \hline
%  			%    $\beta_{u,\bar{u}}$ & A continuous 3D curve segment representing overlap between 3D curves $\Gama_{u}$ and $\Gama_{\bar{u}}$ \\ \hline
%  			$\mathbf A$ & A 3D junction linking 3 or more distinct curve points\\
%  			\hline
%  		\end{tabular}
%  		%\vspace{-0.1cm}
%  	\end{center}
%  %  \vspace{-1cm}
%  \end{table}

Image curve fragments formed from grouped edges are central to our framework.
Each image $V^v$ at view $v = 1,\dots,N$ contains a number of curves
$\gama_i^v$, $i=1,\dots,M^v$. Reconstructed 3D curve fragments are referred
as $\Gama_k$, $k=1,\dots,K$, whose reprojection onto view $v$ is
$\gama^{k,v}$. Indices may be omitted where clear from context.

The initial stage of our framework is built as an extension of the
hypothesize-and-verify 3D Curve Sketch approach~\cite{Fabbri:Kimia:CVPR10}. We
use the same hypothesis generation mechanism
with a novel verification step performing a finer-level
analysis of image evidence and significantly
reducing the fragmentation and redundancy in the 3D models.

Two image curves $\gama^{v_1}_{l_1}$ and $\gama^{v_2}_{l_2}$ are paired from two
distinct views $v_1$ and $v_2$ at a time, the
\emph{hypothesis views}, provided they have sufficient epipolar
overlap~\cite{Fabbri:Kimia:CVPR10}. The verification of these $K$ curve pair
hypotheses, represented as $\omega_k$, $k=1,\dots,K$ with the corresponding 3D
reconstruction denoted as $\Gama_k$, gauges the extent of edge support for the
reprojection $\gama^{k,v}$ of $\Gama_k$ onto another set of \emph{confirmation
views},
$v = v_{i_3},\dots,v_{i_n}$. An image edge in view $v$ suports $\gama^{k,v}$ if it is sufficiently close in
distance \emph{and} orientation. The total support a hypothesis $\omega$ receives from view
$v$ is 
\begin{equation}\label{eq:support}
S^v_{\omega_k} \doteq \int_{0}^{L^{k,v}} \phi(\gama^{k,v}(s))ds,
\end{equation}
where $L^{k,v}$ is the length of $\gama^{k,v}$, and
$\phi(\gama(s))$ is the extent of edge support at
$\gama(s)$. A view is considered a \emph{supporting view}
for $\omega_k$ if $S^v_{\omega_k} > \tau_v$. Evidence from confirmation views is aggregated in
the form
\begin{equation}
\mathcal S_{\omega_k} \doteq \sum_{v = i_3}^{i_n} \left[ S^v_{\omega_k} > \tau_v \right]
S^v_{\omega_k}.
\end{equation}
The set of hypotheses $\omega_k$ whose support $S_{\omega_k}$ exceeds a threshold
are kept and the resulting $\Gama_k$ form the unorganized 3D curves.

Despite these advances, three major shortcomings remain: $(i)$ some 3D curve
fragments are correct for certain portions of the underlying curve and erroneous in other
parts, due to multiview grouping inconsistencies; $(ii)$ 
gaps in the 3D model, typically due to unreliable
reconstructions near epipolar tangencies, where epipolar
lines are nearly tangent to the curves; and $(iii)$ multiple, redundant
3D structures. We now document each issue and describe our solutions.

\begin{figure}
	%xxx \hspace{-0.7cm}
	\begin{center}
		\centering
		\includegraphics[width=\linewidth]{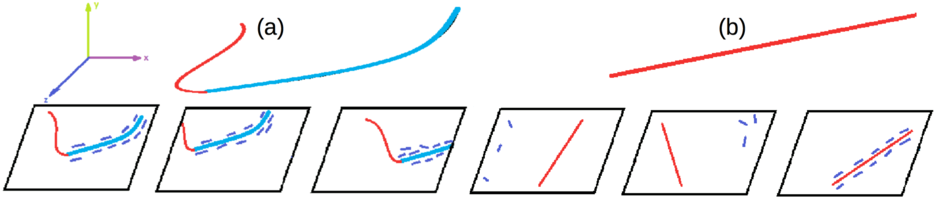}
	\end{center}
	\ReduceBeforeCaptionfigspace
	\caption{\small (a) Due to a lack of consistency in grouping of edges at the image level, a correct 3D curve reconstruction, shown here in blue, can be erroneously grouped with an erroneous reconstruction, shown here in red, leading to
		partially correct reconstructions. When such a 3D curve is projected in its entirety to a number of image views, we only expect the correct portion to gather sustained image evidence, which argues for a hypothesis verification method that can distinguish between supported segments and outlier segments; (b) An incorrect hypothesis can at times 
		coincidentally gather an extremely high degree of support from a limited set of views. The red 3D line shown here might be an erroneous hypothesis, but because parallel linear structures are common in man-made environments, such an incorrect hypothesis often gathers coincidental strong support from a particular view or two. Our hypothesis verification approach is able to handle such cases by requiring explicit support from a minimum number of viewpoints simultaneously.
	}
	\label{fig:sample:localization}
	\ReduceAfterCaptionfigspace
\end{figure}

\begin{problem}
  \textbf{Erroneous grouping:} inconsistent multiview grouping of edges
  can lead to reconstructed curves which are veridical only along some
  portion, which are nevertheless wholly admitted,
  Fig.~\ref{fig:sample:localization}(a). Also, fully-incorrect hypotheses can
  accrue support coincidentally, as with
  repeated patterns or linear structures, Fig.~\ref{fig:sample:localization}(b).  
  Both issues can be addressed by allowing for
  selective local reconstructions: only those portions of the curve receiving adequate
  edge support from sufficient views are reconstructed. This ensures that
  inconsistent 2D groupings do not produce spurious 3D reconstructions. The
  shift from cumulative global to multi-view local support results in greater
  selectivity and deals with coincidental alignment of edges with the
  reconstruction hypotheses.
\end{problem}

%This gives the possibility of deciding
%which portions of a curve can be deemed reliable and which portions not. Specifically, this is done 
%in two ways: a) The constraint that a curve reconstruction receives support
%distributed across multiple views and not just a limited number of views can now be localized to a small portion of 
%the curve, thus producing more accurate results  (see Fig.~\ref{fig:view:spread}) \textcolor{red}{[Anil: is the figure correct?]} and 
%b) The constraint that a curve should receive sufficient support is now localized so 
%some portions are deemed reliable and other are not (see Fig.~\ref{fig:sample:localization}). 

\begin{problem}\textbf{Gaps:}
	The geometric inaccuracy of curve segment reconstructions nearly
	parallel to epipolar lines led~\cite{Fabbri:Kimia:CVPR10} to break off curves at
	epipolar tangencies, creating 2D gaps leading to gaps in 3D.
We observe, however, that while reconstructions near epipolar tangency are
\emph{geometrically unreliable}, they are \emph{topologically correct} in that
they connect the reliable portions correctly but with highly inaccurate
geometry. What is needed is to flag curve segments near epipolar tangency
reconstructions as geometrically unreliable. We do this by the integration of
support in Equation~\ref{eq:support}, giving significantly lower weight to
these unreliable portions instead of fully discarding them, which greatly
reduces the presence of gaps in the resulting reconstruction.
\end{problem}

%\vspace{-0.6cm}

\begin{figure}[h]
	\begin{center}
		\centering
		\includegraphics[width=0.6\linewidth]{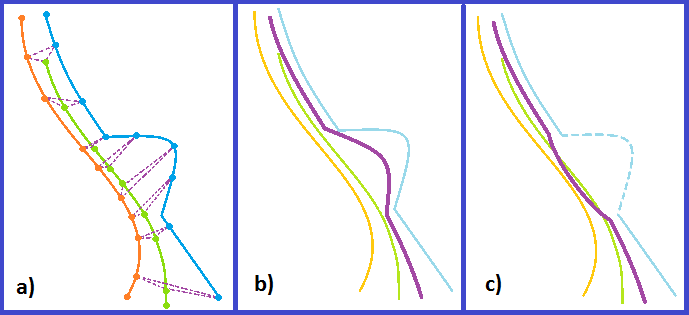}
	\end{center}
	\ReduceBeforeCaptionfigspace
	\caption{\small %
		(a) Redundant 3D curve reconstructions (orange, green and blue) can arise from a single 2D image
		curve in the primary hypothesis view. If the redundant curves are put in
		one-to-one correspondence and averaged, the resulting curve is shown in (b) in purple. Our robust averaging approach, on the other hand, is able to get rid of that bump by eliminating outlier segments, producing the purple curve shown in (c).
	}
	\label{fig:robust:averaging} \ReduceAfterCaptionfigspace
\end{figure}

%\vspace{-0.3cm}

\begin{problem} \textbf{Redundancy:}
  A 2D curve can pair up with dozens of curves from other
  views, all pointing to the same reconstruction, leading to
  redundant pairwise reconstructions as partially overlapping 3D curve segments,
  each localized slightly differently. Our solution is to detect and reconcile
  redundant reconstructions. Since redundancy changes
  as one traverses a 3D curve, we reconcile redundancy at the
  local level: each 3D edge is in one-to-one correspondence with a 2D edge of
  its primary hypothesis view (\ie, the first view from which it was reconstructed),
  hence 3D edges can be grouped in a one-to-one manner, all corresponding to
  a common 3D source. These are robustly averaged by data-driven
  outlier removal, where a Gaussian distribution is fit on all pairwise
  distances between corresponding samples, discarding samples farther than
  $2\sigma$ from the average, Fig.~\ref{fig:robust:averaging}.
  Robust averaging improves localization accuracy, removes redundancy, and
  elongates shorter curve subsegments into longer 3D curves. 
\end{problem}

%\vspace{-0.2cm}

%\vspace{-0.5cm}

\begin{figure}
	\begin{center}
		%\centering
		\includegraphics[height=4cm]{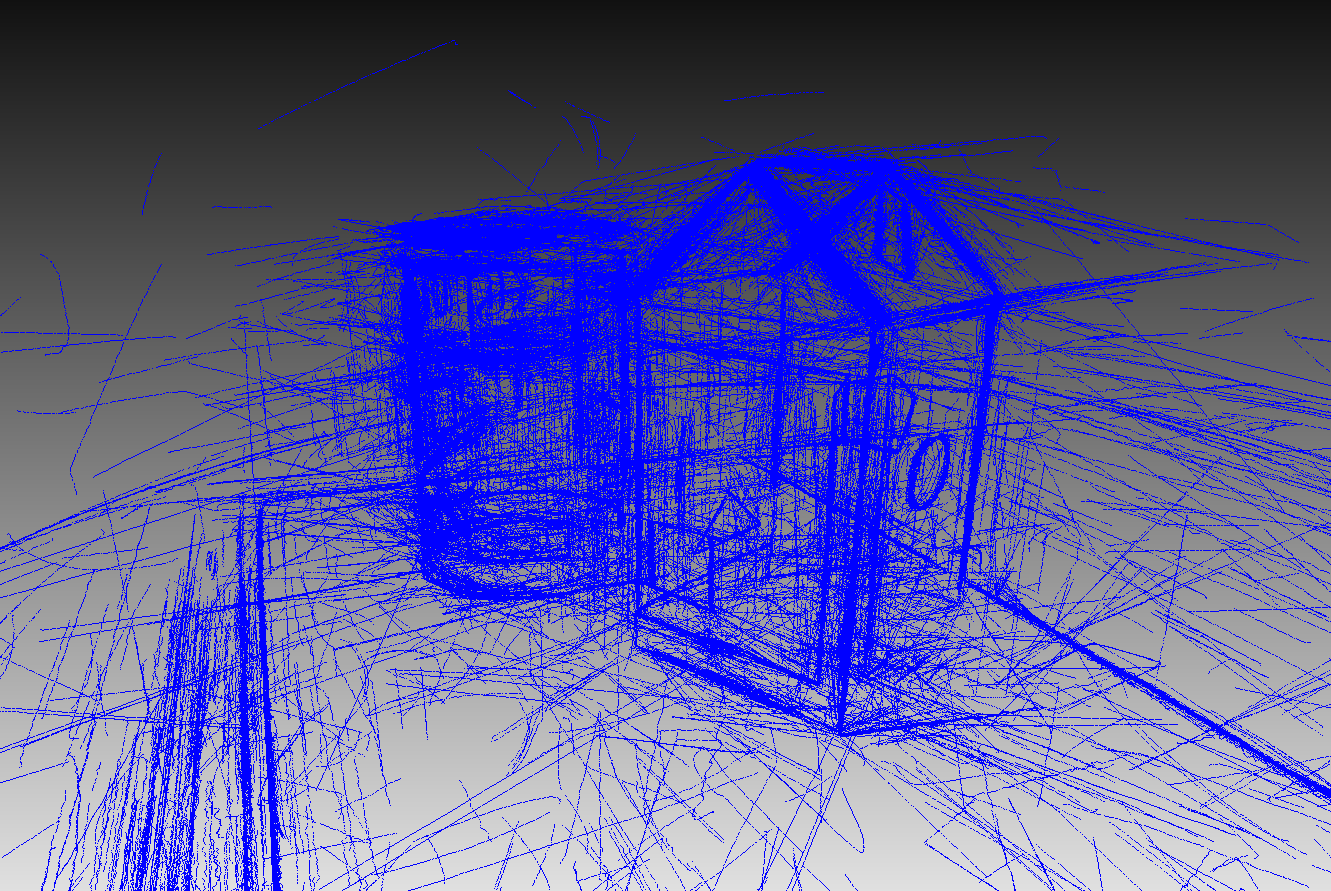}~\includegraphics[height=4cm]{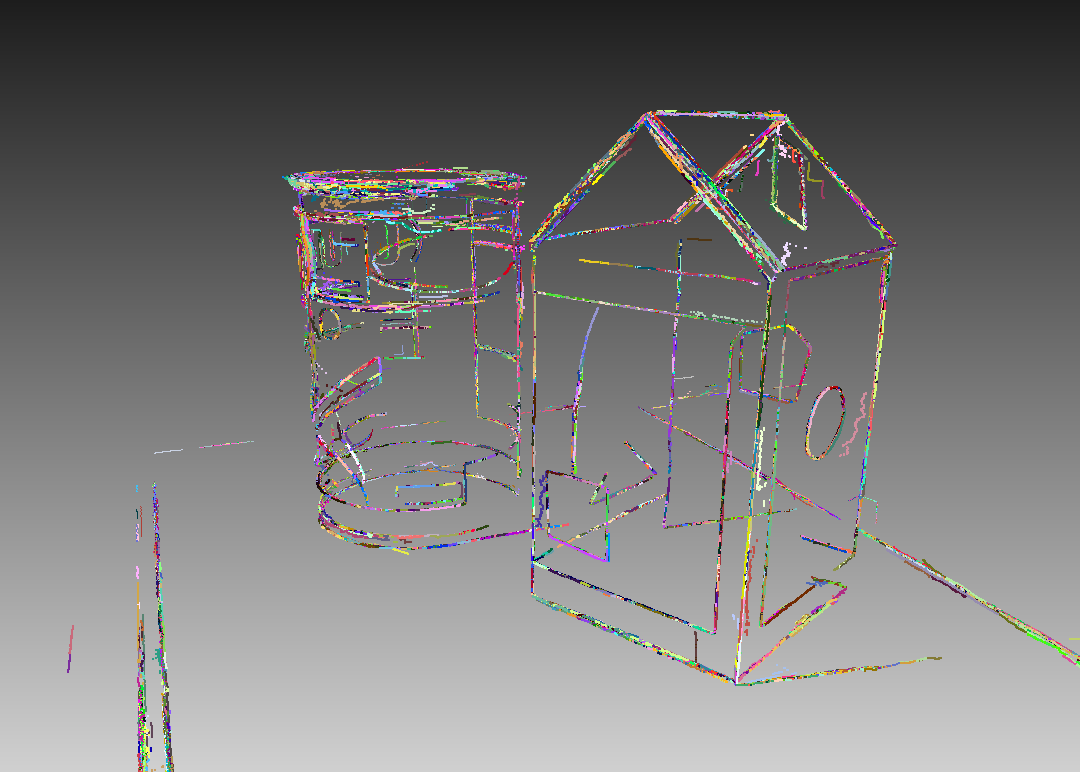}
	\end{center}
	%   \ReduceBeforeCaptionfigspace
%  \vspace{-0.3cm}
	\caption{\small %
		A visual comparison of: (left) the curve sketch
		results~\cite{Fabbri:Kimia:CVPR10}, with (right) the
		results of our enhanced curve sketch algorithm presented in Section~\ref{sec:improved:sketch}.
		Notice the significant reduction in both outliers and duplicate reconstructions, without sacrificing 
		coverage.
	}
	\label{fig:improved:qual} \ReduceAfterCaptionfigspace
\end{figure}

% XXX IMPORTANT rewrite TODO
%Specifically, in the original curve sketch system, multiple instances of hypothesis view pairs
%are unrelated, even for shared hypothesis views. We instead
%perform a robust multiview reconstruction in the latter case.
%Let $\omega_k = \{ \gama_{i_1}^1, \gama_{i_2}^2,\dots \gama_{i_n}^n \}$ be
%a match hypothesis confirmed by the 3D curve sketch (at higher recall than the
%original paper). We reconstruct the underlying $\Gama_k$ and its
%differential geometry (here expressed by the tangent field $\T_k$) by
%simultaneous optimal multiview triangulation of all the matching curve fragments in
%$\omega_k$~\cite{Fabbri:Kimia:EMMCVPR2005,Fabbri:PhD:2010}, instead of pairwise as in the 3D Curve
%Sketch system. Robustness is achieved by outlier rejection.
%\indraftnote{Anil: describe how to do robustness}
%\indraftnote{Ric\& Anil: is there anything worth keeping - Ben.  Yes, after all this is
%the formalization of what was done and is the main contribution of this section! - Ric}

\section{From 3D Curve Sketch to 3D Drawing}
\label{sec:3d:drawing}

Despite the visible improvements of the Enhanced 3D Curve Sketch of
Section~\ref{sec:improved:sketch}, Fig.~\ref{fig:improved:qual},
curves are broken in many places, and there remains redundant overlap.
The sketch representation as unorganized clouds of
3D curves are not able to capture the fine-level geometry or spatial
organization of 3D curves, \eg by using junction points to characterize
proximity and neighborhood relations. The underlying cause of these issues
is lack of integration across multiple views. The robust averaging approach
of Section~\ref{sec:improved:sketch} is one step, anchored on one primary
hypothesis view, but integrates evidence within that view only; a
scene curve can be visible from multiple hypothesis view pairs, and
some redundancy remains.

\begin{figure}[h]%[htpb]
	\centering
	\includegraphics[width=0.49\linewidth]{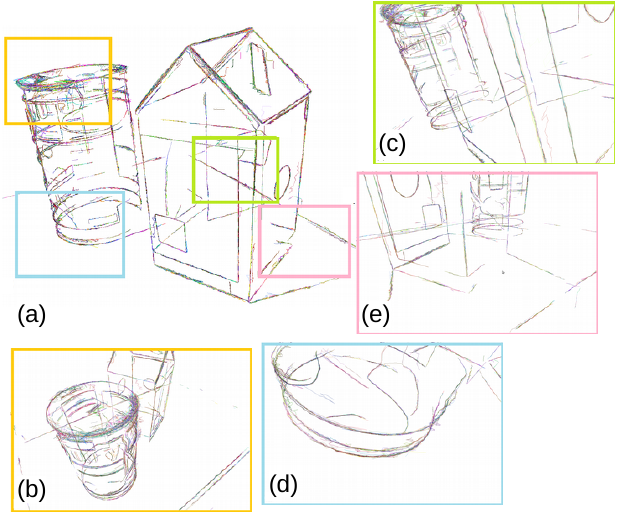}
	\includegraphics[width=0.49\linewidth]{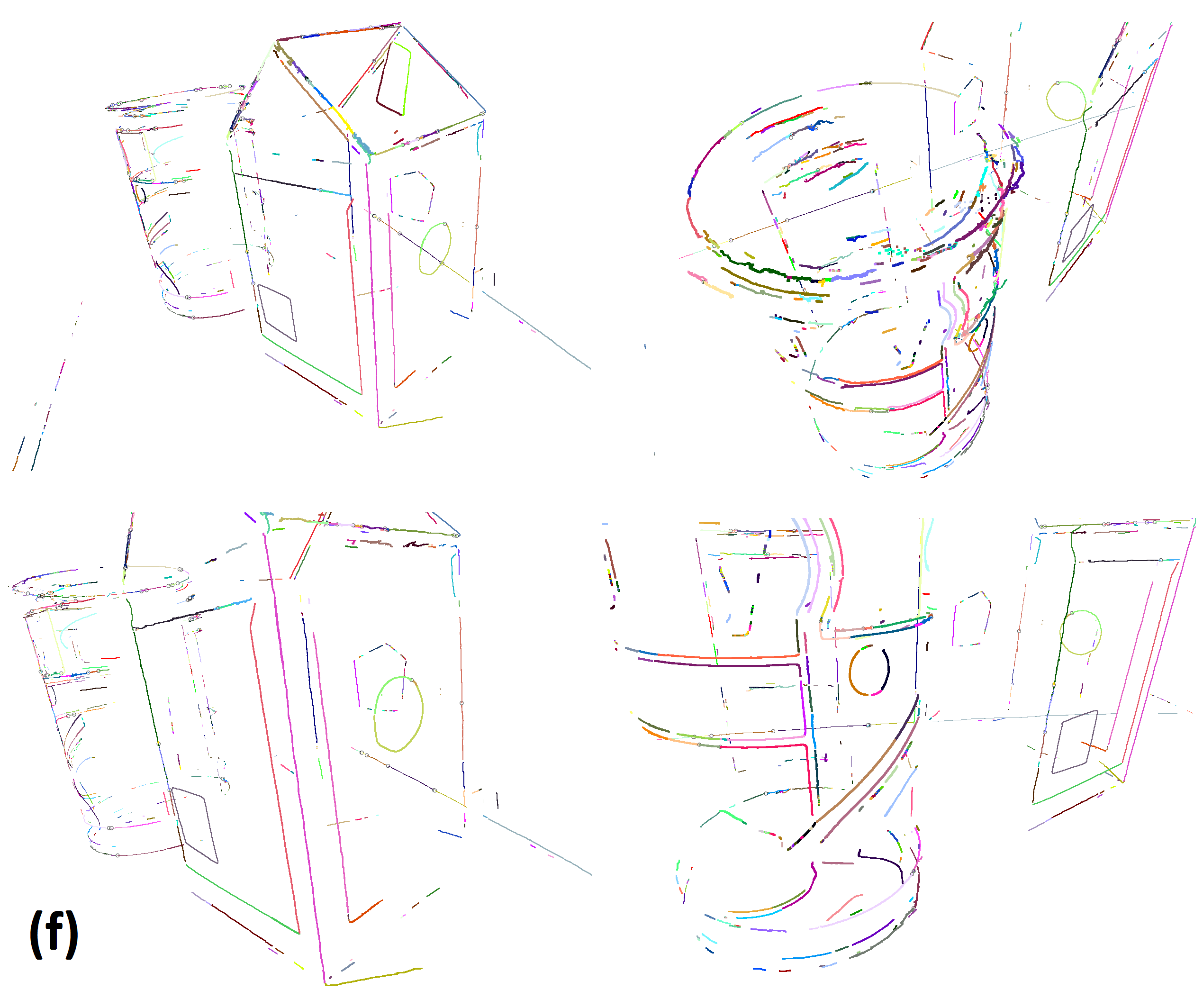}
	%   \includegraphics[width=0.5\linewidth]{figs/allWithROIMarked.png}
	%   \includegraphics[width=0.25\linewidth]{figs/view1FrameMarked.png}
	%   \includegraphics[width=0.25\linewidth]{figs/view2FrameMarked.png}
	%   \includegraphics[width=0.25\linewidth]{figs/view3FrameMarked.png}
	%   \includegraphics[width=0.25\linewidth]{figs/view4FrameMarked.png}
	%\ReduceBeforeCaptionfigspace
%  \vspace{-0.2cm}
	\caption{\small (a) The four main issues with the enhanced curve sketch: (b) localization errors along the camera
		principal axis, which cause loss in accuracy if not corrected, (c) redundant 
		reconstructions due to a lack of integration across different views,
		(d) the reconstruction of a single long curve as multiple,
		disconnected (but perhaps overlapping) short curve segments, and (e) the lack
		of connectivity among distinct 3D curves which naturally form junctions. (f) shows the 3D drawing reconstructed from this enhanced curve sketch, as described in Section~\ref{sec:3d:drawing}. Observe how each of the four bottlenecks have been resolved. Additional results are evaluated visually and quantitatively, and are reported in Section~\ref{sec:results} as well as Supplementary Materials.
	}\label{fig:issues:remaining} 
	\ReduceAfterCaptionfigspace
\end{figure}

This lack of multiview integration is responsible for three
problems observed in the enhanced curve sketch,
Fig.~\ref{fig:issues:remaining}: $(i)$ localization inaccuracies,
Fig.~\ref{fig:issues:remaining}b, due to use of partial information; $(ii)$
reconstruction redundancy, which lends to multiple curves with partial
overlap, all arising from the same 3D structure, but remaining distinct, see
Fig.~\ref{fig:issues:remaining}c; $(iii)$ excessive breaking because each curve
segment arises from one curve in one initial view independently.

\textbf{Multiview Local Consistency Network:}
The key idea underlying integration of reconstructions across views is the
detection of a common image structure supporting two reconstruction hypotheses. 
Two 3D local curve segments depict the same single underlying
3D object feature if they are supported by the same 2D image edge structures. 
Since the identification of common image structure can vary along the curve, it
must necessarily be a local process, operating at the level of a 3D local edge
and not a 3D curve. 
Two 3D edge elements (edgels) depict the same 3D structure if they 
receive support from the same 2D edgels in a sufficient number of views, so
3D-2D links between a 2D edgel to the 3D edgel it supports must be kept. Typically,
they share supporting image edges in many views; and the number of shared
supporting edgels is the measure of strength for a 3D-3D link between them.

Formally, we define the Multiview Local
geometric consistency Network (MLN) as pointwise alignments $\phi_{ij}$ between 
two 3D curves $\Gama_i$ and $\Gama_j$: let $\Gama_i(s_i)$ and $\Gama_j(s_j)$ be
two points in two 3D curves, and define
\begin{equation}
S_{ij} \doteq \{v : \gama^{i,v}(s_i) \text{ and } \gama^{j,v}(s_j) \text{ share local support}\}.
\end{equation}
Then the a kernel function $\phi$ defines a consistency link between these two points,
weighted by the extent of multiview image support $\phi_{ij}(s_i,s_j) \doteq
|S_{ij}|$. When the curves are sampled, $\phi$ becomes an adjacency matrix of a
graph representing links between individual curve samples.
The implementation goes through each image edgel which votes for a 3D curve
point that has received support from it (see the supplementary material for
details).

%\draftnote{cross check terminology w/ amir linking graphs}
%\draftnote{parametrized alignment to write integrals}
%\vspace{-0.3cm}
\begin{figure}
	%\captionsetup[subfigure]{labelformat=empty}
	\centering
	\begin{tabular}{cc}
%    \vspace{-0.3cm}
		\multirow{2}[2]{*}[20mm]{\subfloat[]{\includegraphics[width=0.75\linewidth]{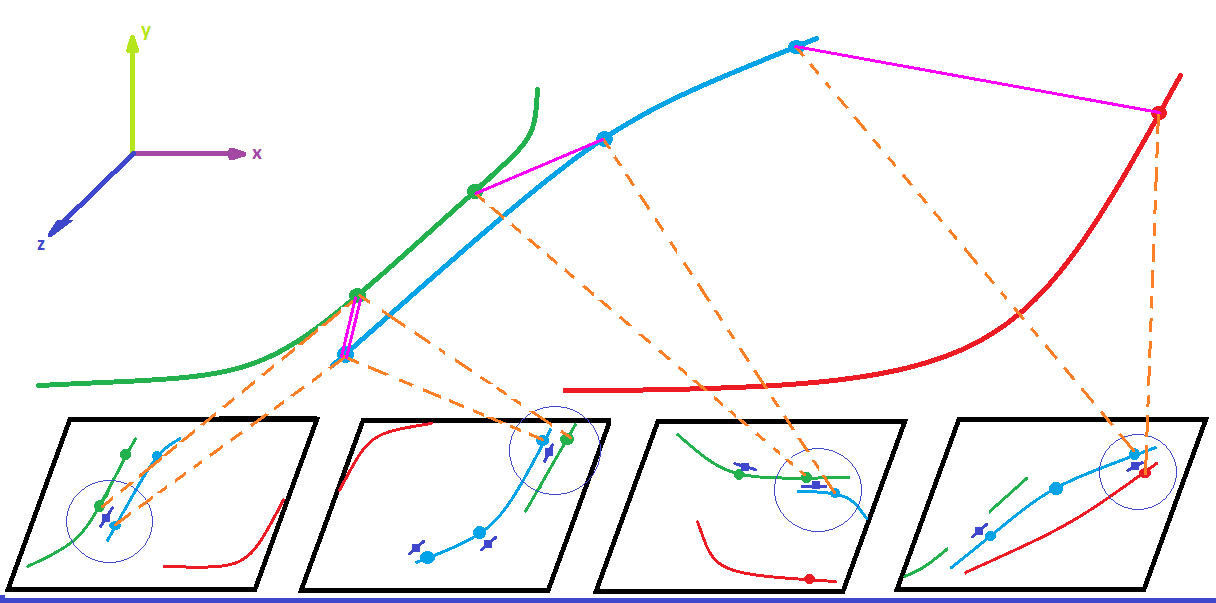}}} &
		\subfloat[]{\includegraphics[width=0.2\linewidth]{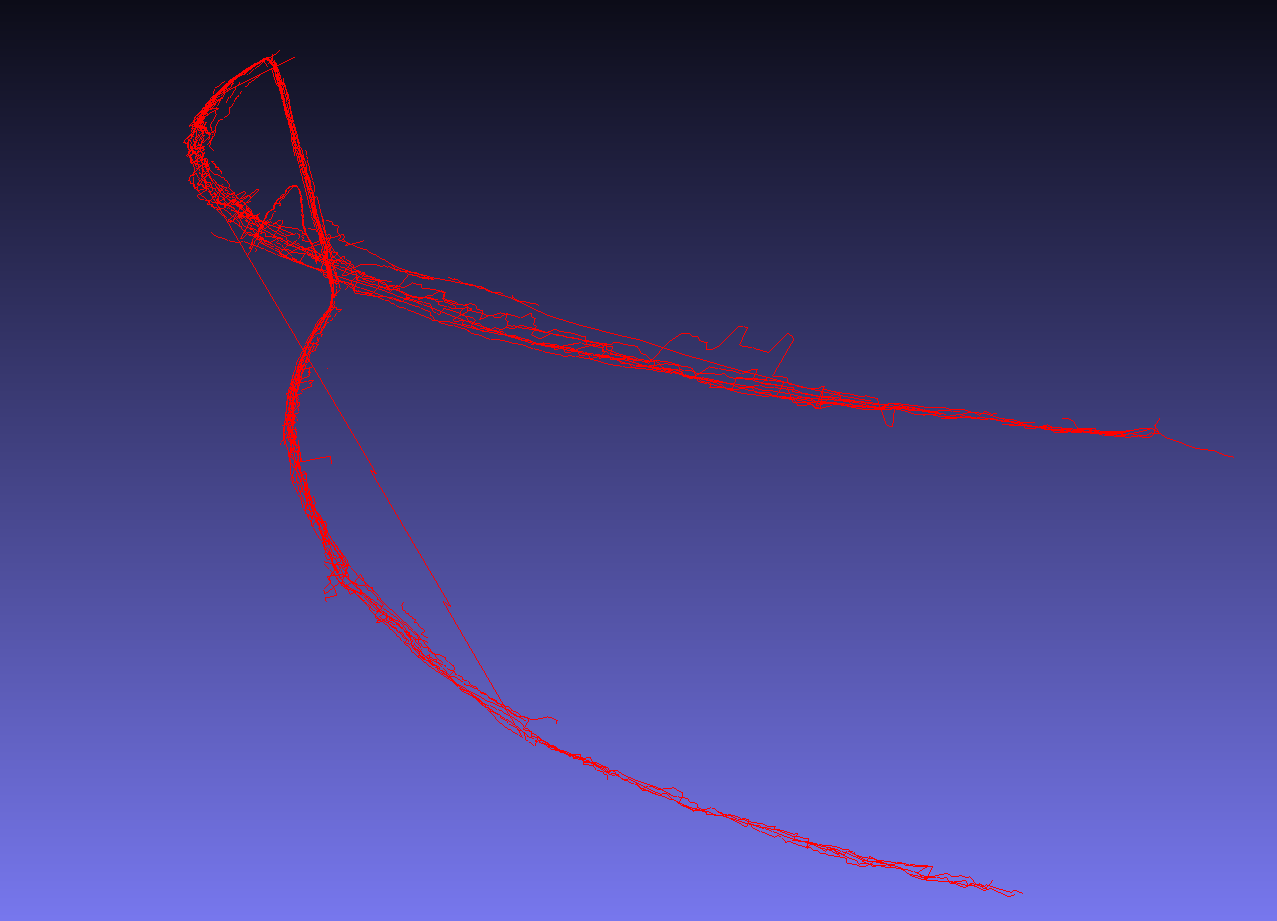}}\\
		&
		\subfloat[]{\includegraphics[width=0.2\linewidth]{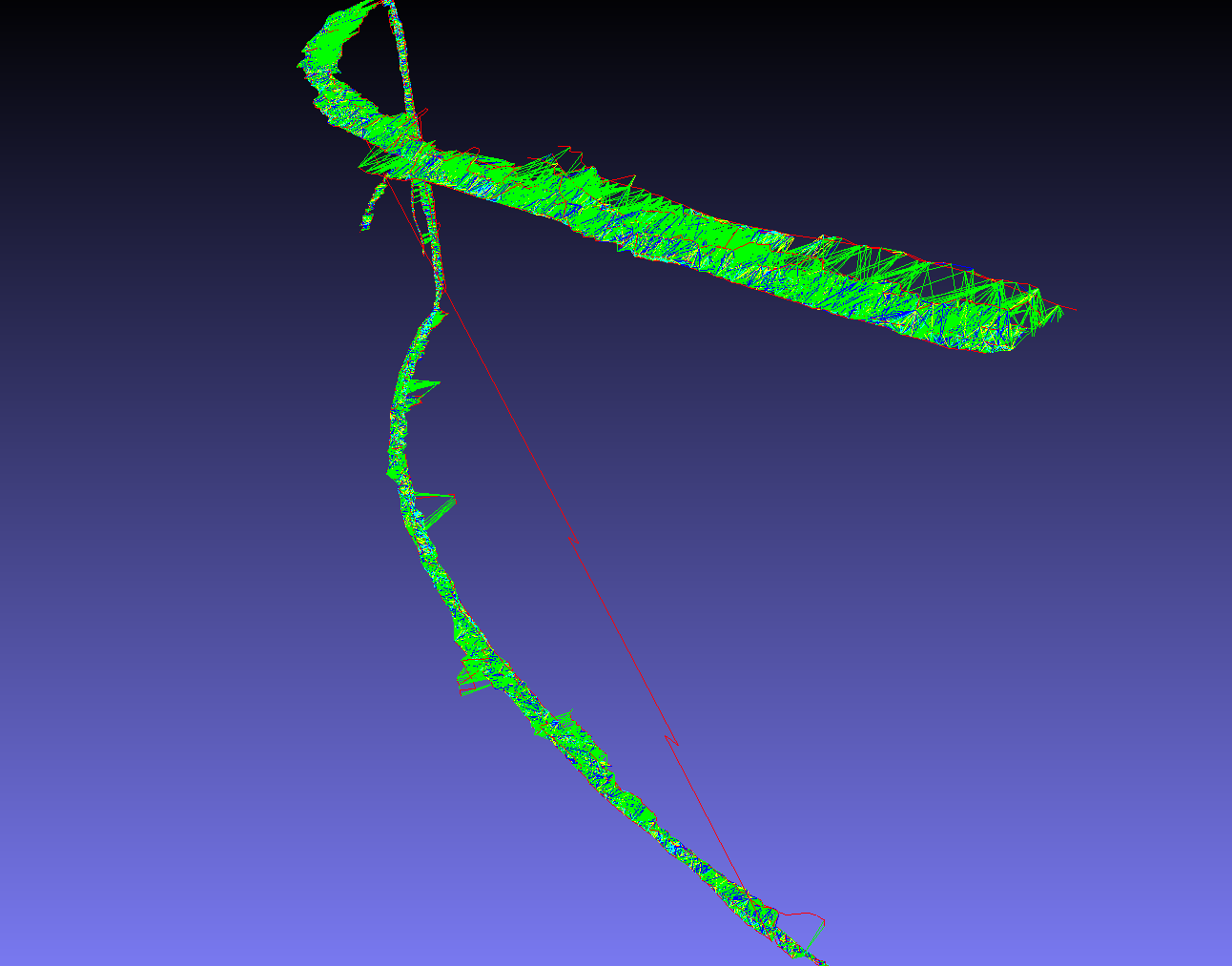}} \\
	\end{tabular}
	%\vspace{-0.3cm}
	\caption{\small The four bottlenecks of
		Fig.~\ref{fig:issues:remaining} are resolved by integration of
		information/cues from all views. (a) The shared supporting edges, which are marked with circles, create the purple links between the corresponding samples of the 3D curves. These purple bonds will then be used to pull the redundant segments together and reorganize the 3D model into a clean 3D graph. Observe how the determination of common	image support can identify portions of the green and blue curves as identical	while differentiating the red one as distinct. A real example for a bundle of related curves is shown in (b) and the links among their edges in (c).
	}
	%   based on which 3D information with a
	%   common cause can be integrated, leading to a four-stage process of $(i)$
	%   establishing the links as pictured above, $(ii)$ identifying portions of 3D
	%   curves which arise from the same source, Fig~\ref{fig:overlap:masks}, $(iii)$
	%   precise localization for 3D curves with a common cause,
	%   Fig~\ref{fig:graph:organization}a, (iv) Topological merging of these
	%   segments, leading to a topological graph of 3D curve segments,
	%   Fig~\ref{fig:graph:organization}e.}\label{fig:master:figure}
	\ReduceAfterCaptionfigspace
\end{figure}

%\vspace{-0.3cm}

\textbf{Multiview Curve-level Consistency Network:}
The identification of 3D edges sharing 2D edges leads to
high recall operating point with many false links due to accidental alignment of
edge support. False positives can be reduced without affecting high
recall by employing a notion of curve context for each 3D edgel: a link
between two 3D edgels based on a supporting 2D edgel is more effective if
the respective neighbors of the underlying 3D edge on the underlying 3D curve
are also linked. 

The curve context idea requires establishing new pairwise links between 3D
curves using MLN, when there are a sufficient number of links with
$\phi_{ij}>\tau_{\epsilon}$ between their constituent 3D edges (in our
implementation, $\tau_{\epsilon}=3$ and we require 5 such edges or more). The
linking of 3D curves is represented by the Multiview Curve-level Consistency
network (MCCN), a graph whose nodes are the 3D curves $\Gama_j$ and the edges
represent the presence of high-weight 3D edge links between these 3D curves. The
\mccn graph allows for a clustering of 3D curves by finding connected
components; and once a link is established between two curves, there is a high
likelihood of their edges corresponding in a regularized fashion, thus
fewer common supporting 2D edges are required to establish a link between all
their constituent 3D edges. This fact is used to perform gap filling, since even
no edge support is acceptable to fill in small gaps and create a continuous and
regularized correspondence if both neighbors of the gap are connected (see
pseudocode in Supplementary Materials for details). The two stages in
tandem, \ie, high recall linking of 3D edges and use of curve context to reduce
false positives leads to high recall and high precision, \textit{i.e.}, all the
3D edges which need to be related are related and very few outlier connections
remain.

\begin{figure}
	\captionsetup[subfigure]{labelformat=empty}
	\centering
	\begin{tabular}{cc}
		\multirow{2}[2]{*}[13mm]{\includegraphics[width=0.55\linewidth]{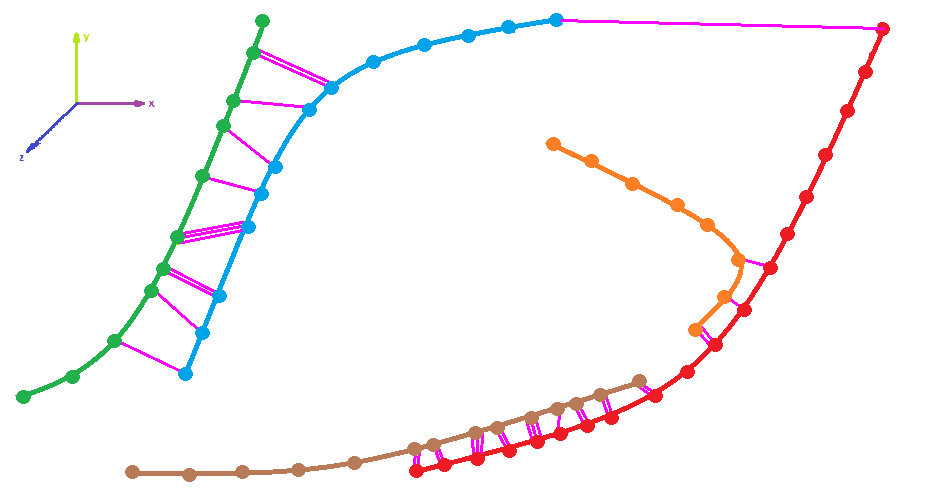}} &
		\includegraphics[width=0.45\linewidth]{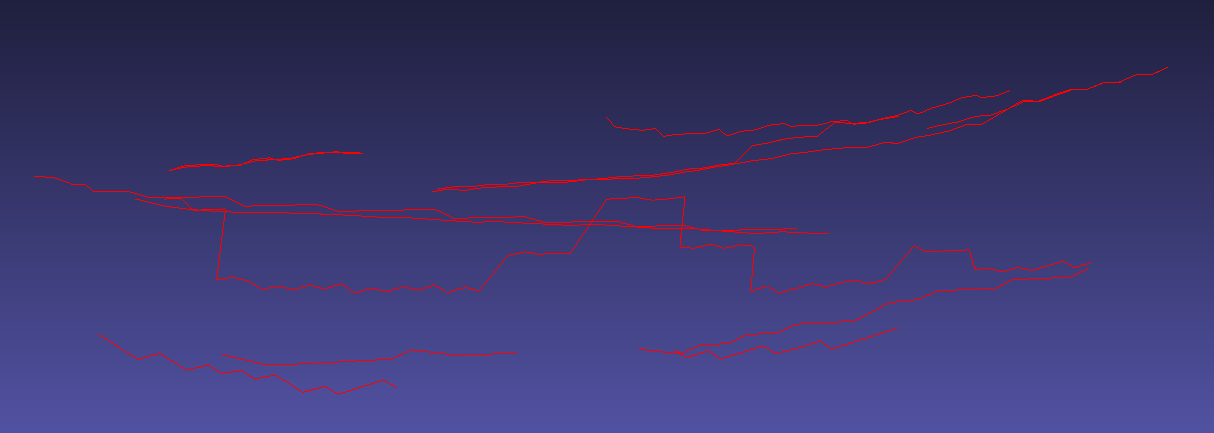}\\
		&
		\includegraphics[width=0.45\linewidth]{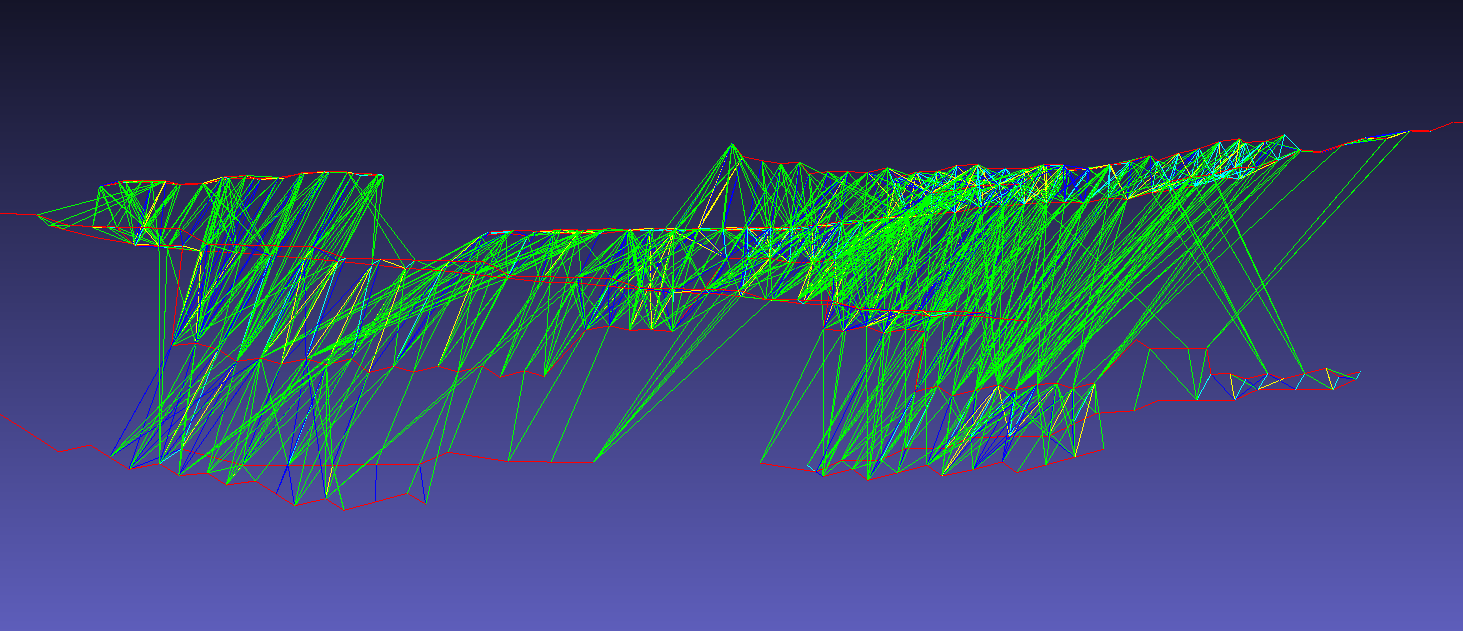}
		\\
	\end{tabular}
	%\ReduceBeforeCaptionfigspace
	\caption{\small 
		The correspondence between 3D edge samples is skewed along a curve, a direct indication that these links cannot be used as-is when averaging and fusing redundant curve reconstructions. Instead, each point is assumed to be in correspondence with the point closest to it on another overlapping curve, during the iterative averaging step. Observe that corrections can be partial along related
		curves.
	}
	\label{fig:overlap:masks}
	\ReduceAfterCaptionfigspace
\end{figure}

%\vspace{-0.3cm}

%The links between two 3D curves that are consistently overlapping according to
%evidence from multiple views form the so-called Multiview Curve-level
%Consistency Network, a graph defined as follows.
%\begin{definition}
%The Multiview Curve-level Consistency Network (MCCN) is a graph 
%\begin{equation}
%\text{\textit{MCCN}} = (S, L),
%\end{equation}
%where the vertices $S = \{1,\dots,K\}$ encode the set of 3D curves
%$\{\Gama_1,\dots,\Gama_K\}$ and $L$ is the link set defined below.
%\end{definition}
%\begin{definition}
%Let the set of so-called strong local links between curves $\Gama_i$ and
%$\Gama_j$ be defined as
%\begin{equation}
%S_{ij} \doteq \{(s,t) : \phi_{ij}(s,t) \geq \tau_{s}, \phi_{ij} \in
%\text{MLN}(\Gama_1,\dots,\Gama_K) \}.
%\end{equation}
%Then the set $L$ of the MCCN is defined as
%\begin{equation}
%L = \{(i,j) : |S_{ij}| \geq \tau_{sl}\}.
%\end{equation}
%\end{definition}
%\indraftnote{We can formulate L as an integral over alignment parameter, see
%rics notes}

%We cluster the 3D curves by computing the connected components of their MCCN.
%We then go back to local correspondences by filling-in short correspondence gaps
%of the MLN induced by the confidence gained by the curve-level consistency link.
%Further details are available in the form of a pseudocode in supplementary material.

\begin{figure}[htbp]
	\begin{center}
		\centering
		\includegraphics[width=0.86\linewidth]{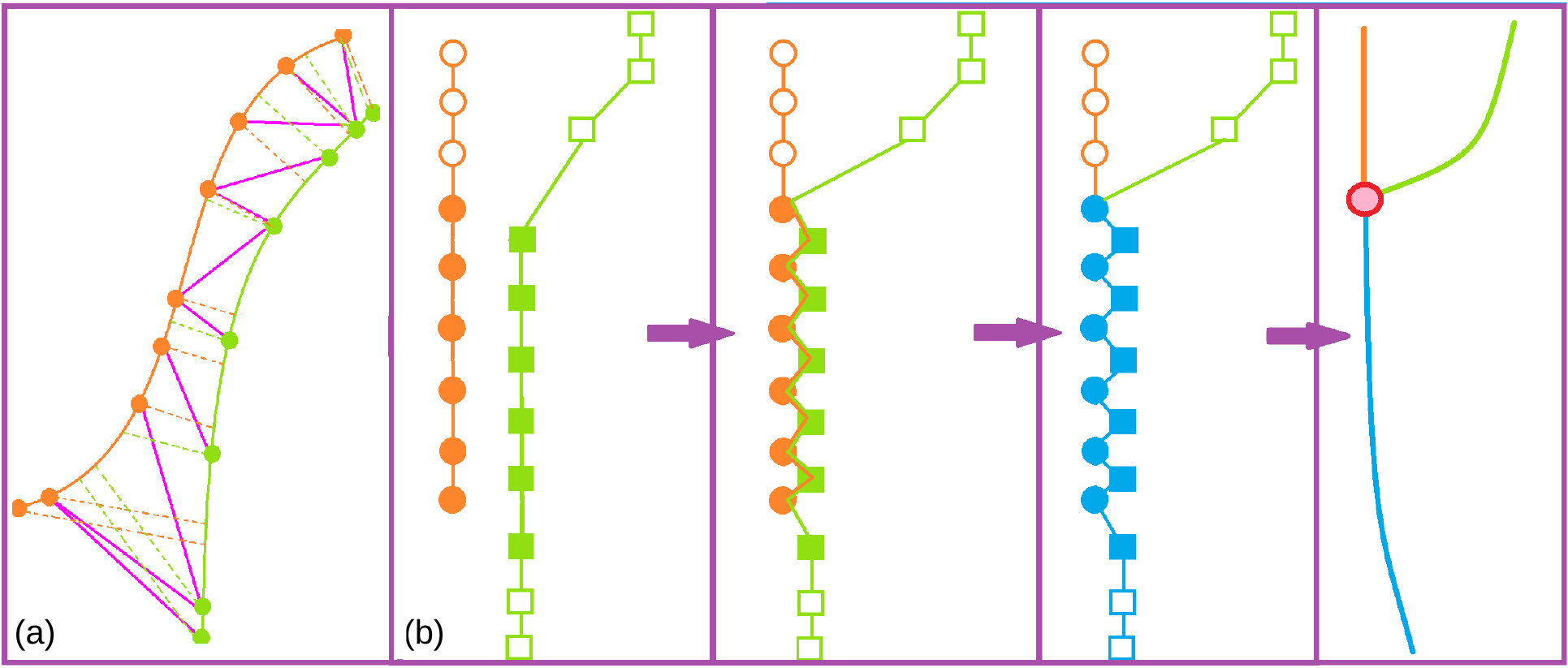}
		\includegraphics[height=2.8cm]{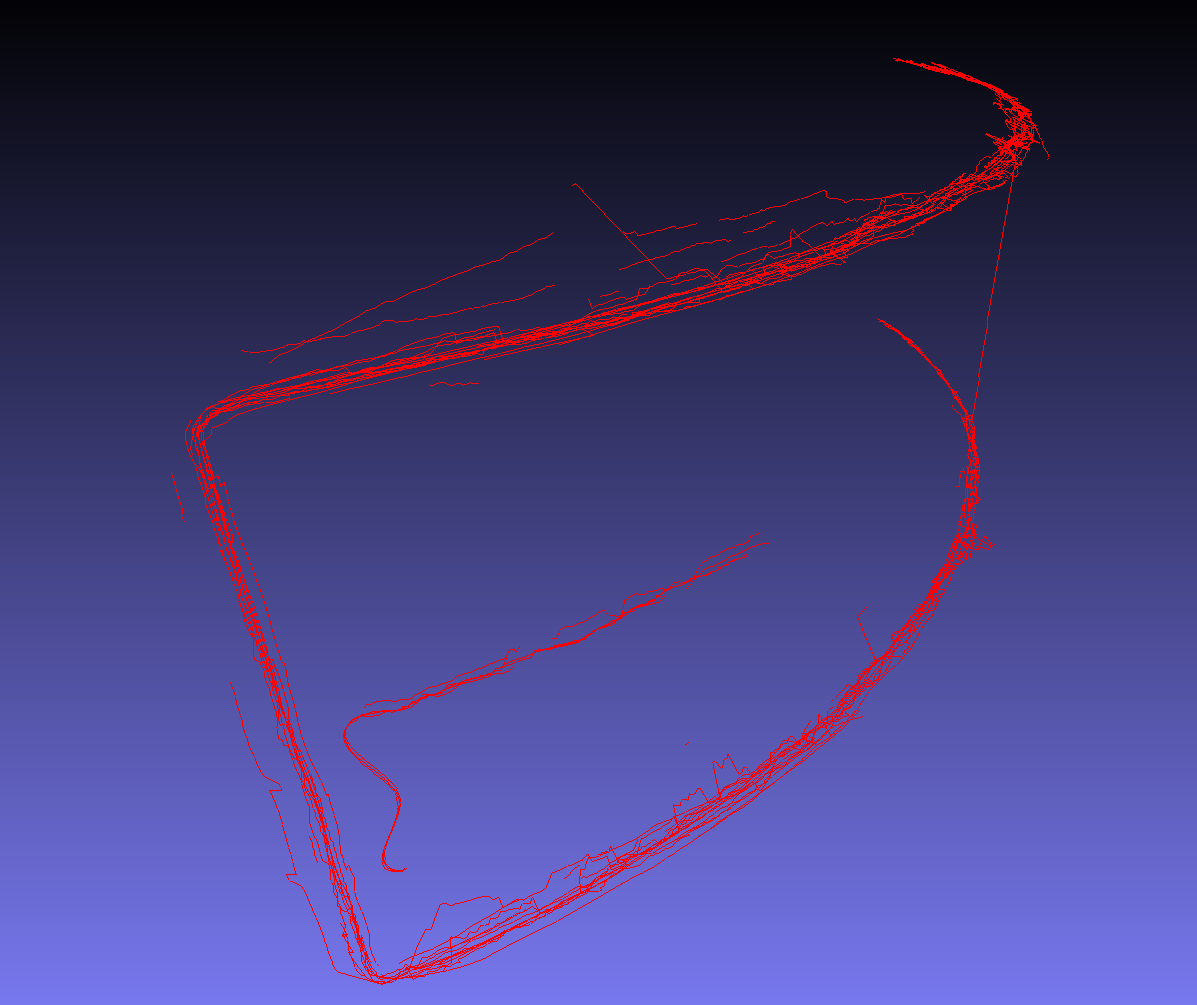}
		\includegraphics[height=2.8cm]{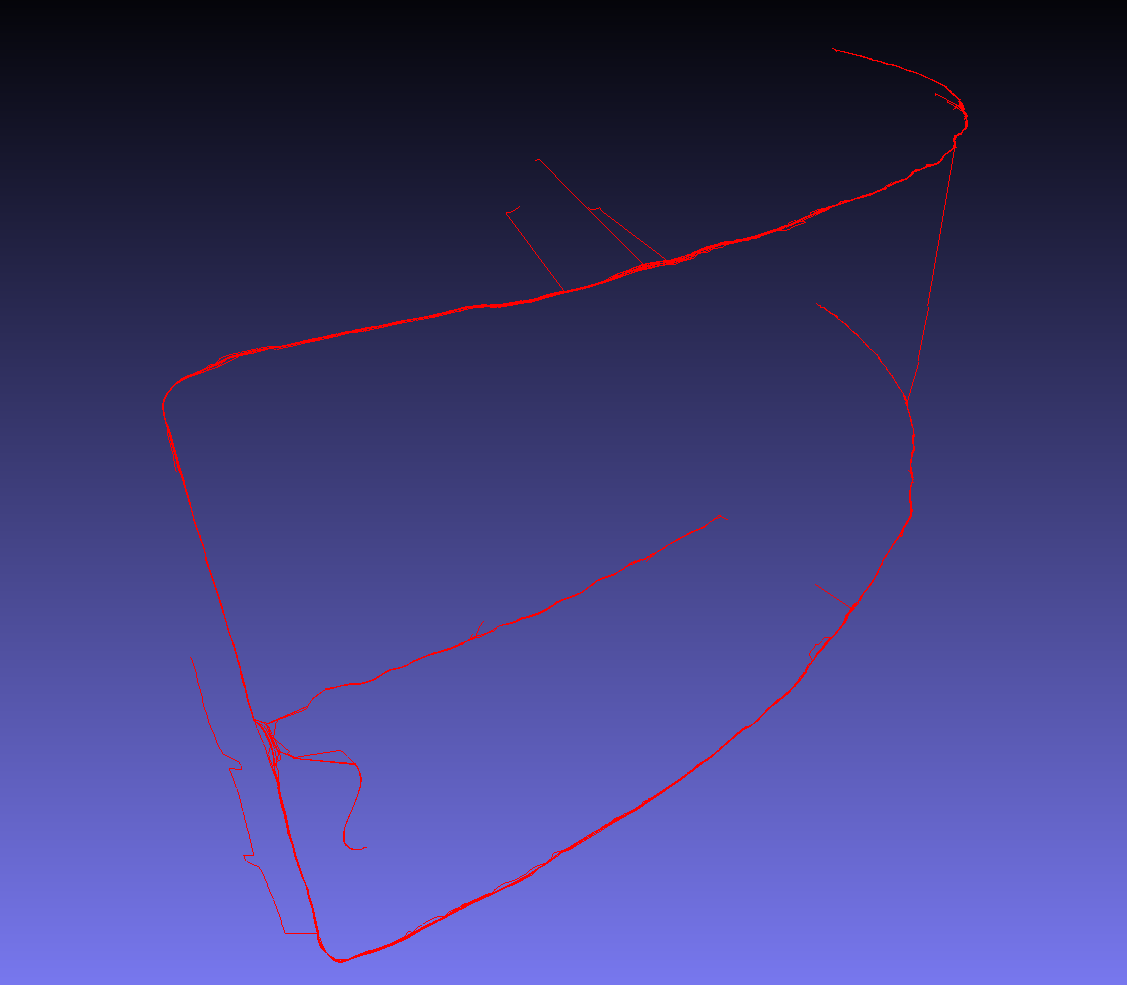}
		\includegraphics[height=2.8cm]{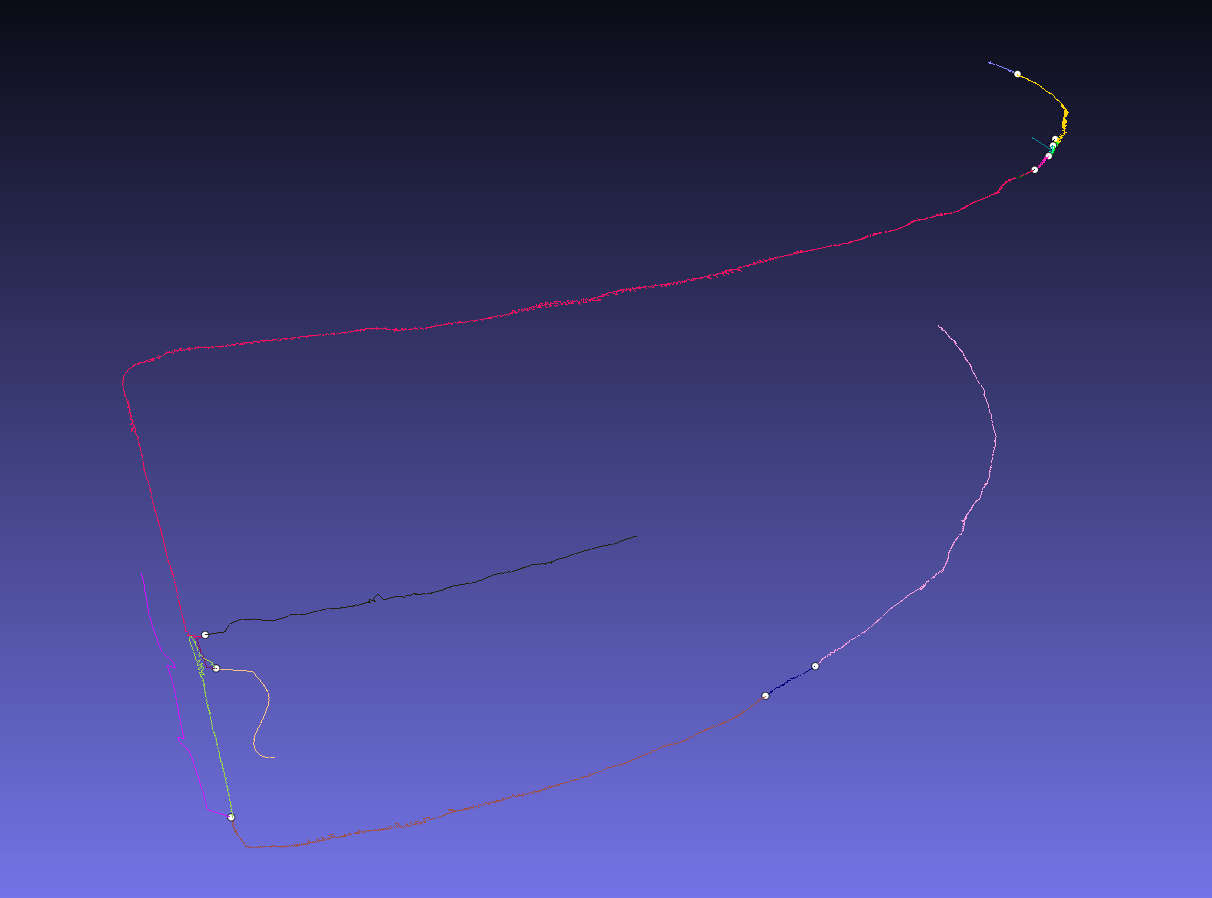}
	\end{center}
	\ReduceBeforeCaptionfigspace
	\caption{\small (a) A schematic of sample correspondence along two related 3D
		curves, showing skewed correspondences that may not be one-to-one. (b) A
		sketch of how two curves are integrated. Bottom row: a real case.
	}
	\label{fig:graph:organization}
	\ReduceAfterCaptionfigspace
\end{figure}

\textbf{Integrating information across related edges:} 
The identification of a bundle of curves as arising from the same 3D source
implies that we can improve the geometric accuracy of this bundle by
allowing them to converge to a common solution. While this might appear
straightforward, 3D edges are not consistenly distributed along related curves,
yielding a skew in the correspondence of related samples,
Fig.~\ref{fig:overlap:masks}, sometimes not a one-to-one
correspondence, Fig.~\ref{fig:graph:organization}a. This argues for averaging 3D
curves and not 3D edge samples, which in turn requires finding a more
regularized alignment between the 3D curves, without gaps; we find each
curve samples's closest point on the other curve.

When post averaging a sample with its closest points on related curves, the
order of resulting averaged samples is not clear.  The order should be inferred
from the underlying curves, but this information can be
conflicting, unless the distance between two curves is substantially smaller
than the sampling distance along the curves. This requires
first updating each curve's geometry separately and iteratively, without
merging curves until after convergence,
Fig.~\ref{fig:graph:organization}d. This also improves the correspondence of
samples at each iteration, as the closest points are continuously updated. 

At each stage, the iterative averaging process simply replaces each 3D edge
sample with the average of all closest points on curves related to it,
Fig~\ref{fig:graph:organization}b--d. This can be formulated
as evolving all 3D curves by averaging along the \mccn using closest points.
Formally, each $\Gama_i$ is evolved according to
%{\small
\begin{equation}
\frac{\partial\Gama_i}{\partial t}(s) = \alpha 
\avg_{\substack{(i,j)\in L\\ (\Gama_i, \Gama_j) \in \text{ \textsc{mccn}}}}
\{\Gama_j(r) : \Gama_j(r) = \text{cp}_j(\Gama_i(s)) \},
\end{equation}%}
where $\text{cp}_i(\mathbf p)$ is the closest point in $\Gama_i$ to $\mathbf p$ and $L$ is the link set defined as follows:
Let the set $S_{ij}$ of so-called strong local links between curves $\Gama_i$ and
$\Gama_j$ be
\begin{equation}
S_{ij} \doteq \{(s,t) : \phi_{ij}(s,t) \geq \tau_{\epsilon}, \phi_{ij} \in
\text{MLN}(\Gama_1,\dots,\Gama_K) \}.
\end{equation}
Then the set $L$ of the \mccn is defined as
\begin{equation}
L \doteq \{(i,j) : |S_{ij}| \geq \tau_{sl}\}.
\end{equation}
In practice, the averaging is robust and $\alpha$ is chosen such that in
one step we move to the average.

\textbf{3D Curve Drawing Graph:}
Once all related curves have converged, they can be merged into single curves,
separated by junctions where 3 or more curves meet. The order along the
resulting curve is also dictated by closest points: The immediate neighbors of
any averaged 3D edge are the two closest 3D edges to it among all
converged 3D edges in a given \mccn cluster.

\begin{figure}[h]
	\begin{center}
		\centering
		\includegraphics[width=0.6\linewidth]{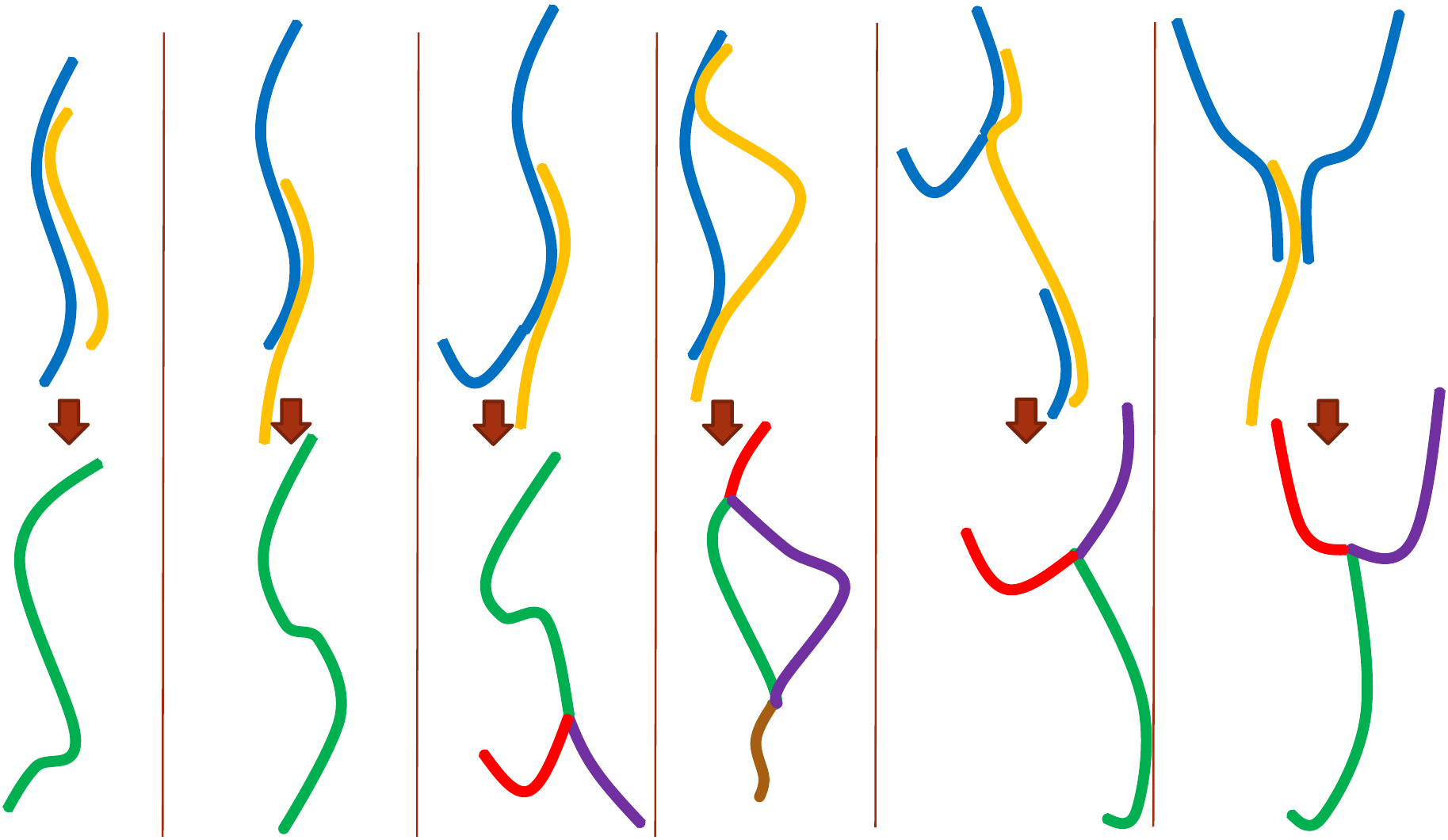}
	\end{center}
	\ReduceBeforeCaptionfigspace
	\caption{\small The complete set of merging primitives, which were
  systematically worked out to cover all possible merging topologies between a
pair of curves whose overlap regions are computedbeforehand. We claim that any configuration of overlap between two curves can be broken down into a series of these primitives along the length of one of the curves. The 5th primitive is representative of a “bridge” situation, where the connection at either end of the yellow curve can be any one of the first four cases shown, and 6th primitive is representative of a situation where only one end of the yellow curve connects to multiple existing curves, but not necessarily just two.}
	\label{fig:junction:topology}
	\ReduceAfterCaptionfigspace
\end{figure}

This where junctions naturally arise: as two distinct curves may merge along one
portion they may diverge at one point, leaving two remaining, non-related
subsegments behind, Fig~\ref{fig:graph:organization}e.  This is a \emph{junction
node} relating three or more curve segments, and its detection is done using the
\emph{merging primitives}, whose complete set are shown in
Fig.~\ref{fig:junction:topology}. The intuition is this: a
complex merging problem along the full length of two 3D curves actually consists
of smaller, simpler and independent merging operations between different
segments of each curve. A full merging problem between two complete
curves can be expressed as a permutation of any number of simpler merging
primitives. These primitives were worked out systematically to serve as the
basic building blocks capable of constructing \emph{all possible configurations}
of our merging problem. 

%in this way, the set of these primitives is not an
%arbitrary heuristic but rather is a complete set of basic operations capable of
%breaking any merging problem into smaller chunks.

%\vspace{-0.8cm}

After iterative averaging, all resulting curves in any given cluster
are processed in a pairwise fashion using these primitives: initialize the 3D
graph with the longest curve in the cluster, and merge every curve in
the cluster one by one into this graph. At each step, any number of these
merging primitives arise and are handled appropriately. This process outputs the
Multiview Curve Drawing Graph (MDG), which consists of multiple disconnected 3D
graphs, one for each 3D curve cluster in the MCCN. The nodes of each graph are
the junctions (with curve endpoints) and the links are curve fragment
geometries. This structure is the final 3D curve drawing.

%\vspace{-0.5cm}

\section{Experiments and Evaluation}\label{sec:results}

We have devised a number of large real and synthetic multiview datasets,
available at \url{multiview-3d-drawing.sourceforge.net}.

% xxx how many points?

%\vspace{-0.7cm}

\begin{figure}
	\captionsetup[subfigure]{labelformat=empty}
	\centering
	\begin{tabular}{ccc}
		\vspace{-4mm}
		\multirow{2}[2]{*}[10mm]{\includegraphics[width=0.57\linewidth]{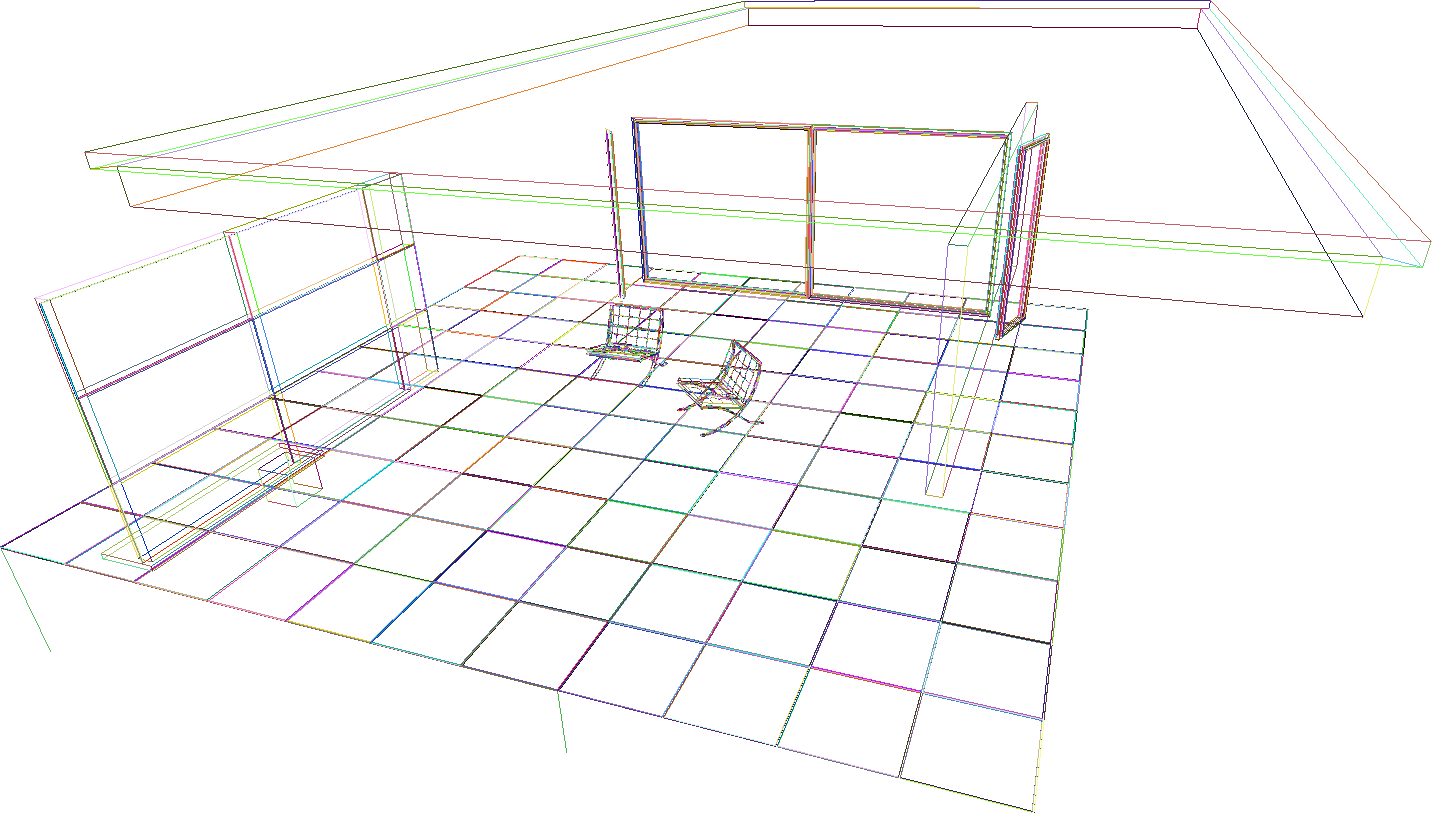}} &
		\subfloat{\includegraphics[width=0.2\linewidth]{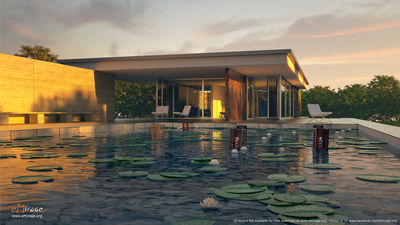}} &
		\subfloat{\includegraphics[width=0.2\linewidth]{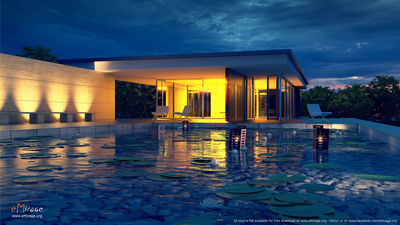}}
		\\
		&
		\subfloat{\includegraphics[height=2.86cm,width=0.2\linewidth]{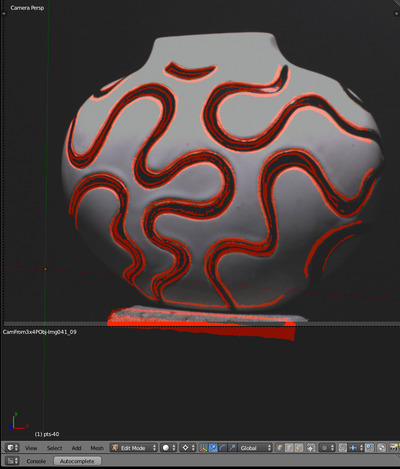}} &
		\subfloat{\includegraphics[height=2.86cm,width=0.2\linewidth]{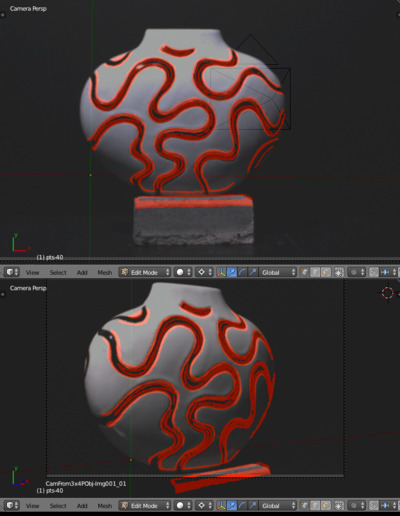}}
		\\
	\end{tabular}
	%\ReduceBeforeCaptionfigspace
%  \vspace{-0.3cm}
	\caption{\small %
    Our publicly-available synthetic (left and top-right) and \textbf{real}
    structured lighting (bottom-right) 3D ground truths modeled and rendered
		using Blender for the present work.
	}
	\label{fig:synth:gt} \ReduceAfterCaptionfigspace
\end{figure}

%\vspace{-0.3cm}
\textbf{The Barcelona Pavilion Dataset:} a realistic synthetic dataset
we created for validating the present approach with control over
illumination, geometry and cameras.  It consists of:
3D models composing a large, mostly man-made, scene professionally composed
by eMirage studios using the 3D modeling
software Blender; ground-truth cameras fly-by's around
chairs with varied reflectance models and cluttered background; $(iii)$
ground-truth videos realistically rendered with high
quality ray tracing under 3 extreme illumination
conditions (morning, afternoon, and night); $(iv)$ ground-truth 3D
\emph{curve} geometry obtained by manually tracing over the meshes.
This is the first synthetic 3D ground truth for evaluating multiview reconstruction algorithms
that is realistically complex -- most existing ground truth is
obtained using either laser or structured light methods, both
of which suffer from reconstruction inaccuracies and calibration errors.
Starting from an existing 3D model ensures that our ground truth is not polluted
by any such errors, since both 3D model and the calibration parameters are
obtained from the 3D modeling software, Fig.~\ref{fig:synth:gt}.
The result is the first
publicly available, high-precision 3D curve ground truth dataset to be used in
the evaluation of curve-based multiview stereo algorithms. For the experiments
reported in the main manuscript we use 25 views out
of 100 from this dataset, evenly distributed around the primary objects of
interest, namely the two chairs, see Fig.~\ref{fig:synth:gt}.

%\vspace{-0.3cm}
\textbf{The Vase Dataset:} constructed for this research from the
\textsc{dtu} Point Feature Dataset with 
calibration and 3D ground truth from structured light~\cite{Aan:Pedersen:etal:IJCV2012,Jensen:etal:CVPR14}.
The images were taken using an automated robot arm from pre-calibrated positions
and our test sequence was constructed using views from different illumination
conditions to simulate varying illumination. To the best of our knowledge, these
are the most exhaustive public multiview ground truth datasets. To generate
ground-truth for curves, we have constructed a GUI based on Blender to manually
remove all points of the ground-truth 3D point-cloud that correspond to
homogeneous scene structures as observed when projected on all views,
Fig.~\ref{fig:synth:gt}(bottom). What remains is a dense 3D point cloud ground truth
where the points are restricted to be near abrupt intensity
changes on the object, {\em i.e.} edges and curves. Our results on this real dataset
showcase our algorithm's robustness under varying illumination. 

%\vspace{-0.3cm}
\textbf{The Amsterdam House Dataset:} 50 calibrated multiview
images, also developed for this research, comprising a wide variety of object properties,
including but not limited to smooth surfaces, shiny surfaces, specific
close-curve geometries, text, texture, clutter and cast shadows,
Fig.~\ref{fig:recon:results}.  The camera reprojection
error obtained by Bundler~\cite{Argarwal:Snavely:etal:ICCV09} is on average subpixel.
There is no ground truth 3D geometry for this dataset; the intent here is: 
to qualitatively test on a scene that is challenging to approaches that rely on,
\eg, point features; and to be able to closely inspect expected geometries
and junction arising from simple, known shapes of scene objects.

%\vspace{-0.3cm}
\textbf{The Capitol High Building:} 256 HD frames from a high $270^\circ$ helicopter fly-by of
the Rhode Island State Capitol~\cite{Fabbri:Kimia:CVPR10}.
Camera parameters are from the Matlab Calibration toolbox and tracking 30
points.
%\vspace{-0.5cm}

\begin{figure}[t]%[h]%[htpb]
  \hspace{-0.2cm}\includegraphics[height=3.2cm]{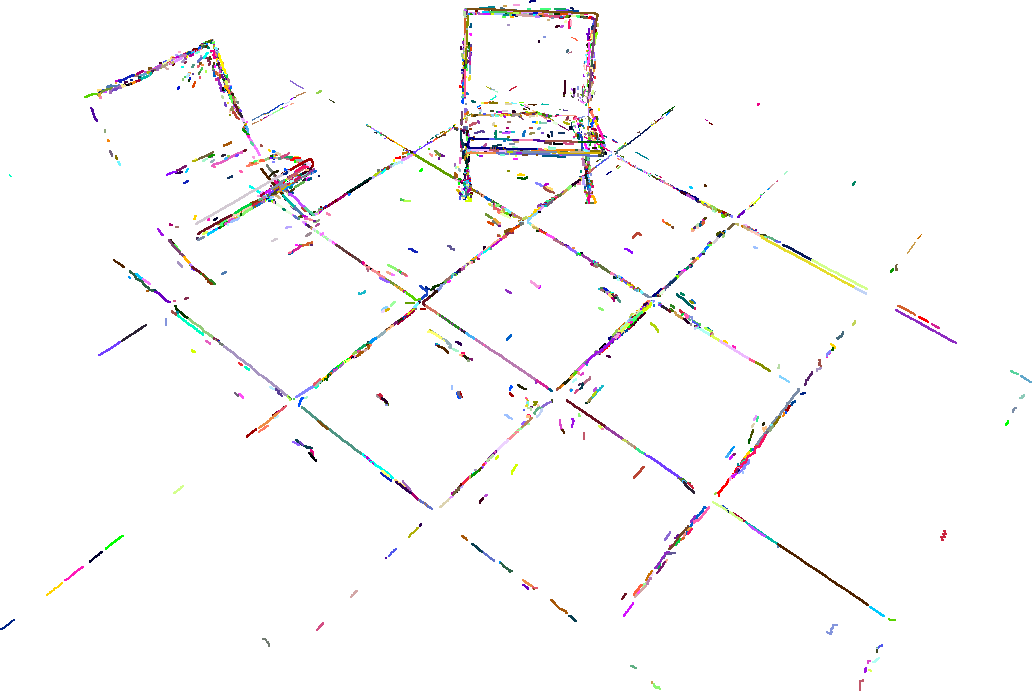}~\hspace{0.2cm}~\includegraphics[height=3.0cm]{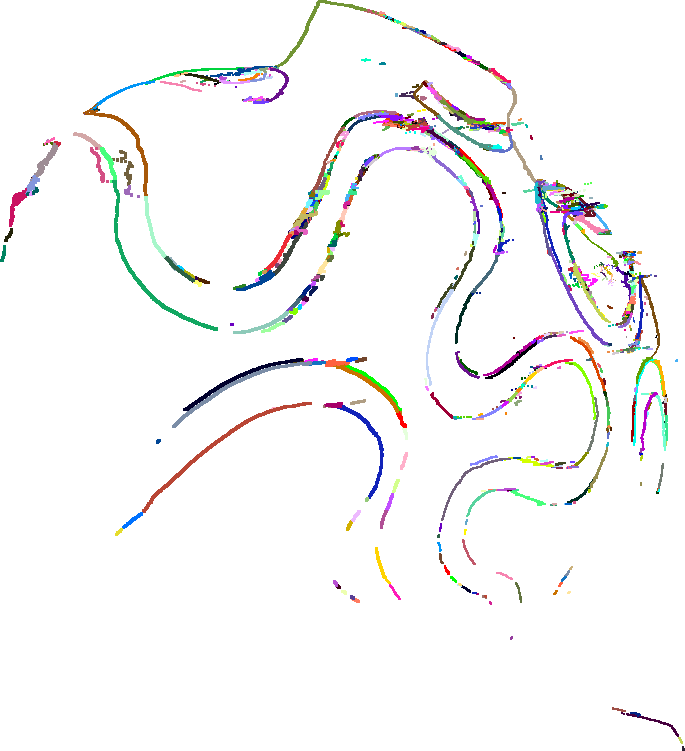}~\hspace{0.2cm}~\includegraphics[height=3.0cm]{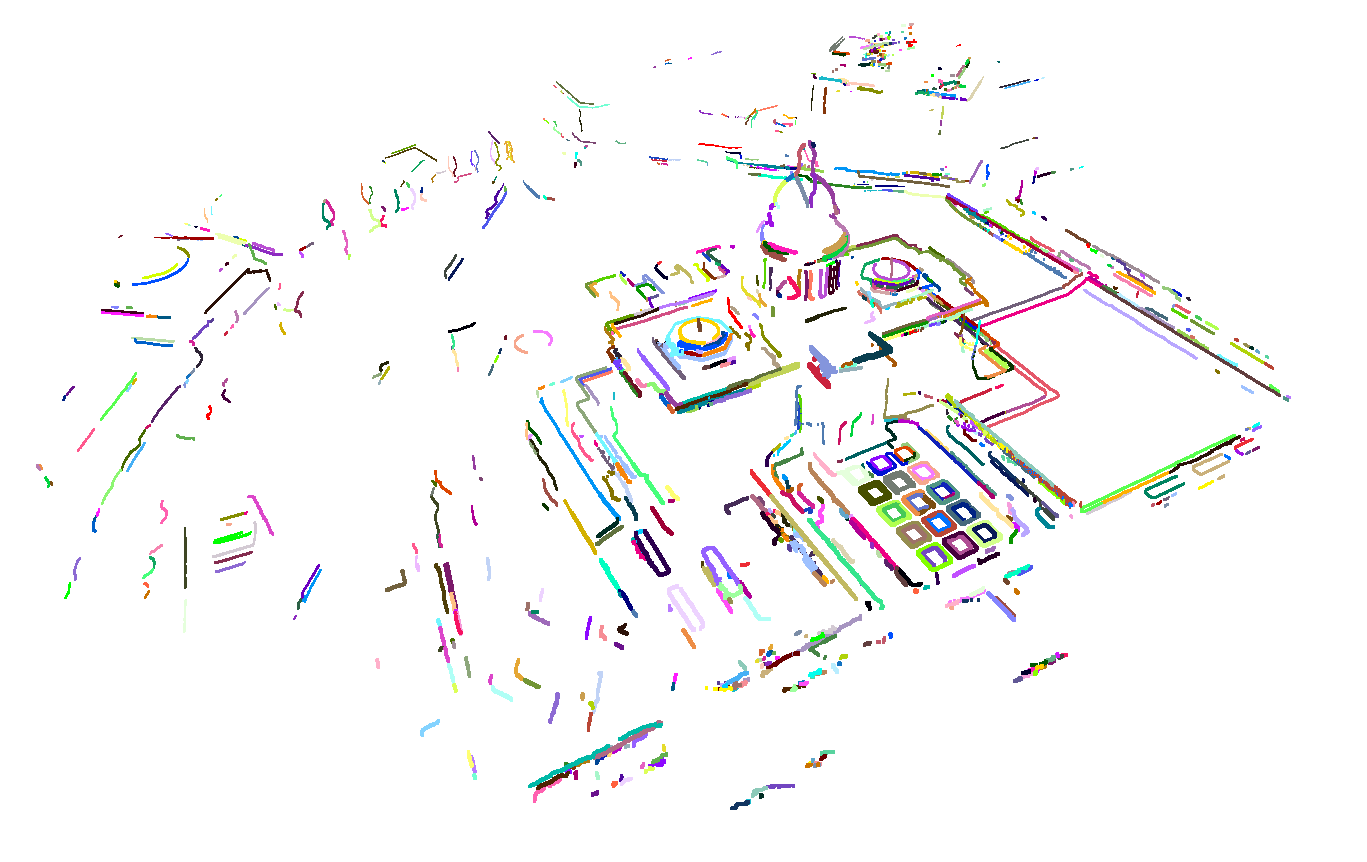}
	\caption{\small The 3D drawing results on the Barcelona Pavilion, DTU Vase and Capitol Datasets. See Supplementary Materials for more extensive results and comparisons
	}\label{fig:pavilion:barcelona:capitol} 
	\ReduceAfterCaptionfigspace
\end{figure}

%\vspace{-1em}
\textbf{Qualitative Evaluation:} The enhancements of
Section~\ref{sec:improved:sketch} lead to significant improvements to the 3D
curve sketch of~\cite{Fabbri:Kimia:CVPR10} in increasing recall while
maintaining precision.  See Fig.~\ref{fig:improved:qual} for a qualitative
comparison.  When the clean clouds of curves are
organized into a set of connected 3D graphs, the results are more accurate, more
visually pleasing and not redundant, Fig.~\ref{fig:issues:remaining}(f)
and Fig.~\ref{fig:pavilion:barcelona:capitol}. Each of the issues in
Fig.~\ref{fig:issues:remaining}(a-e) have been resolved and spatial organization
of 3D curves have been captured as junctions, represented by small
white spheres.

%Another advantage of the enhanced curve sketch is compression: it reduces redundant constructions to represent the same
%scene with less number of curves and curve samples. Fig.~\ref{fig:compression:enhanced} plots the number of curve 
%samples for various scenes that we have reconstructed using both algorithms.
%\vspace{-1em}
%\vspace{-0.7cm}
\begin{figure}
	\centering
	\includegraphics[width=0.325\linewidth]{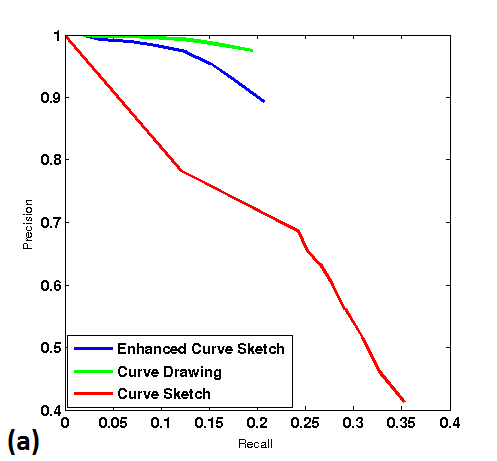}
	\includegraphics[width=0.325\linewidth]{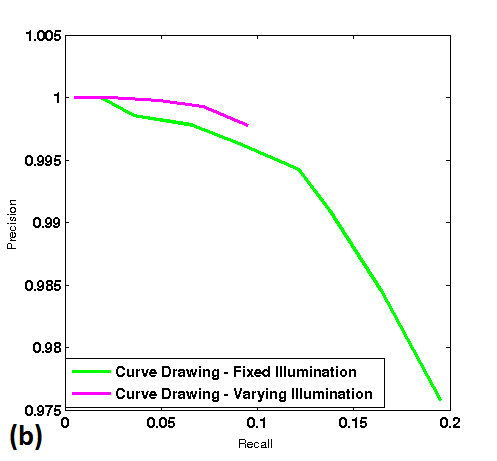}
	\includegraphics[width=0.325\linewidth]{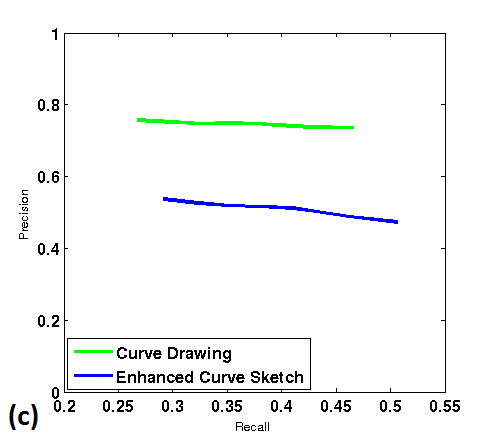}
	\caption{\small %
		Precision-recall curves for quantitative evaluation of 3D curve drawing algorithm: (a) Curve sketch, enhanced curve sketch and curve drawing results are compared on Barcelona Pavilion dataset with afternoon rendering, showing significant improvements in reconstruction quality; (b) A comparison of 3D curve drawing results on fixed and varying illumination version of Barcelona Pavilion dataset proves that 3D drawing quality does not get adversely affected by varying illumination;
		(c) 3D drawing improves reconstruction quality by a large margin in Vase dataset, which consists of images of a real object under slight illumination variation.}
	\label{fig:drawing:quan} \ReduceAfterCaptionfigspace
\end{figure}
%\afterpage{\clearpage}
%\vspace{-0.8cm}
\textbf{Quantitative Evaluation:} 
Accuracy and coverage of 3D curve reconstructions is evaluated against ground
truth. We compare 3 different results to quantify our improvements: $(i)$
Original Curve Sketch~\cite{Fabbri:Kimia:CVPR10} run exhaustively on all views,
$(ii)$ Enhanced Curve Sketch, Section~\ref{sec:improved:sketch}, and $(iii)$
Curve Drawing, Section~\ref{sec:3d:drawing}.  Edge maps are obtained using
Third-Order Color Edge Detector~\cite{Tamrakar:Kimia:ICCV07}, and are linked
using Symbolic Linker~\cite{Yuliang:etal:CVPR14} to extract curve fragments for
each view. Edge support
thresholds are varied during reconstruction for each method, to obtain precision-recall curves. Here, {\bf precision} is the
percentage of accurately reconstructed curve samples: a ground truth curve
sample is a true positive if its closer than a proximity threshold to the
reconstructed 3D model. A reconstructed 3D sample is deemed a
false positive if its not closer than $\tau_{prox}$ to any ground truth curve.
This method ensures that redundant reconstructions aren't rewarded multiple times.  All remaining curve samples in the
reconstruction are false positives. {\bf Recall} is the fraction of ground truth
curve samples covered by the reconstruction. A ground truth sample is marked as
a false negative if its farther than $\tau_{prox}$ to the test reconstruction.
The precision-recall curves shown in Fig.~\ref{fig:drawing:quan}
\emph{quantitatively} measure the improvements of our algorithm and showcase
its robustness under varying illumination.

\section{Conclusion}
%\vspace{-0.3cm}
We have presented a method to extract a 3D drawing as a graph of 3D curve
fragments to represent a scene from a large number of multiview imagery. The 3D
drawing is able to pick up contours of objects with homogeneous surfaces where
feature and intensity based correlation methods fail. The 3D drawing can act as
a scaffold to complement and assist existing feature and intensity based
methods. Since image curves are generally invariant to image transformations
such as illumination changes, the 3D drawing is stable under such changes. The
approach does not require controlled acquisition, does not restrict the number
of objects or object properties.

\clearpage
%\newpage
\appendix
\begin{center}
\huge{\textbf{Supplementary Material}}\\
\large{\textbf{A Brief Summary}}\\
Full supplementary material, code and datasets available at \url{multiview-3d-drawing.sourceforge.net}
\end{center}

This appendix presents additional descriptions, details and results that could
not go into the main paper due to space constraints. This is a subset of the
full supplementary materials available at
\url{multiview-3d-drawing.sourceforge.net}. Section~\ref{sec:more:dataset}
discusses the 3D ground truth benchmarks that were used in the quantitative
evaluation of our results, and details the process with which 3D curvilinear
ground truth models were obtained with the aid of Blender, for both synthetic
and real data. In Section~\ref{sec:moreresults} we present additional figures
for our results.

\section{Obtaining Ground Truth for Quantitative Evaluation}\label{sec:more:dataset}

Quantitative evaluation of 3D models reconstructed from a sequence of images is a non-trivial task due to the difficulties involved in obtaining clean and accurate ground truth 3D models for physical objects in the world, as well as precise calibration for each of the images in the sequence. The well-known Middlebury benchmark~\cite{Seitz:etal:CVPR06} evaluates full surface reconstructions, and the ground truth 3D models are not made public; therefore it is not possible to appropriate them for quantitative evaluation of curve reconstructions. The EPFL benchmark~\cite{Strecha:etal:CVPR08} makes the ground truth 3D models publicly available, but these datasets are limited in the number of views in the image sequence, as well object types and illumination conditions captured in the scene. In our case, the difficulty is compounded by the fact that our reconstruction is a wireframe representation, whereas almost all existing ground truth for multiview stereo is for evaluating dense surface reconstruction algorithms.

\begin{figure}
	\vspace{-0.3em}
	\centering
	\includegraphics[width=0.993\linewidth]{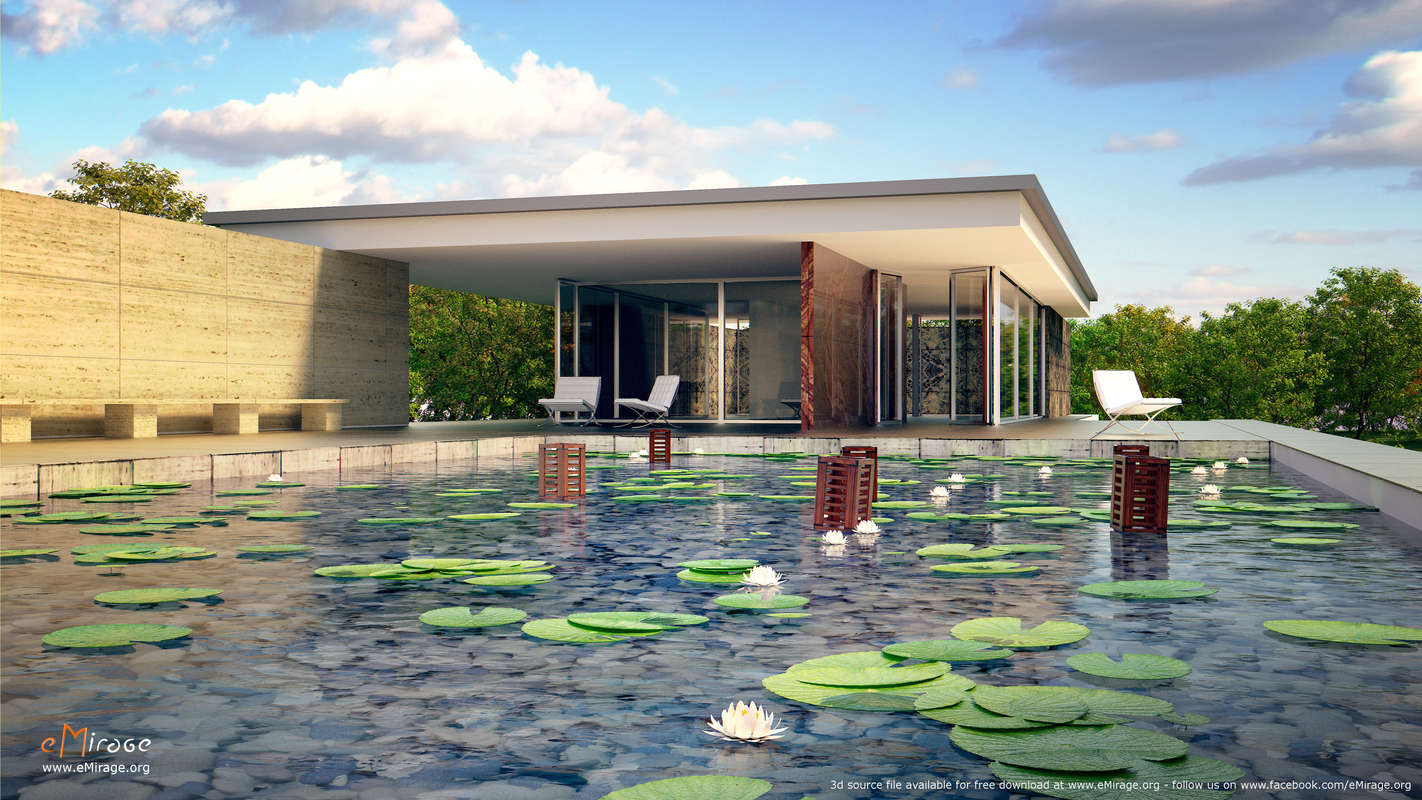}\\
	\includegraphics[width=0.495\linewidth]{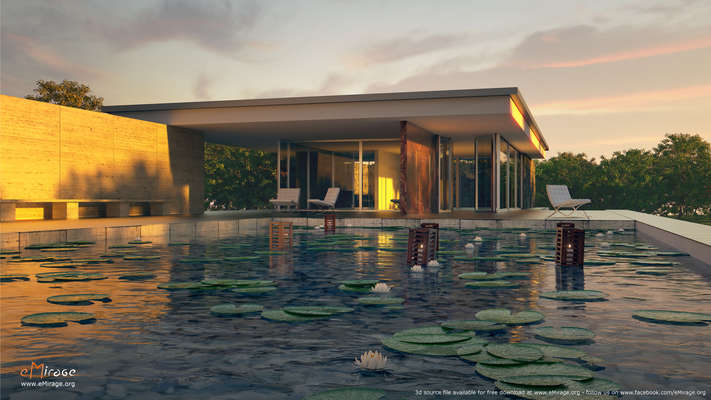}\hspace{0.2mm}%
	\includegraphics[width=0.495\linewidth]{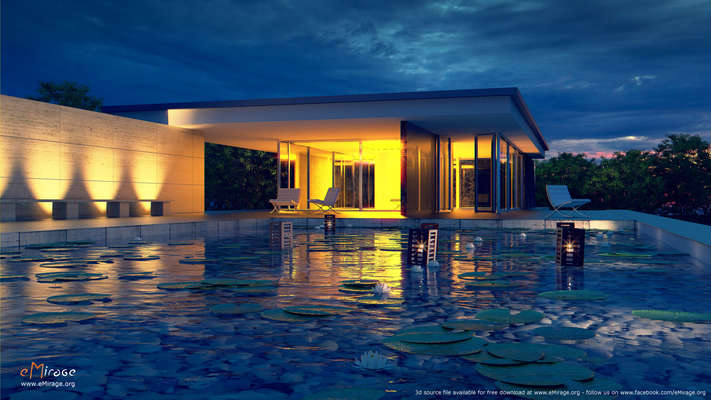}\\[1mm]
	\hspace{-0.3cm}\includegraphics[width=1.015\linewidth]{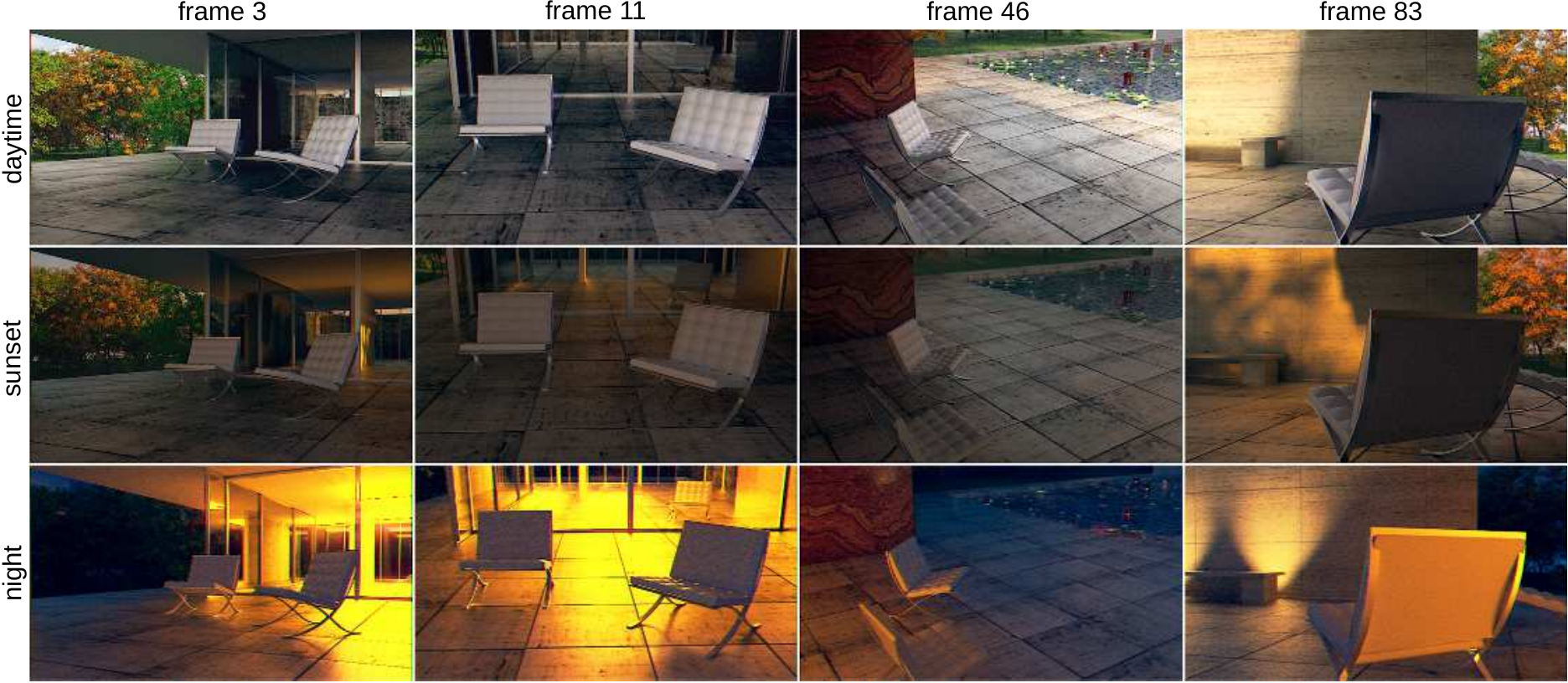}
	\ReduceBeforeCaptionfigspace
	%\vspace{-1em}
	\caption{\footnotesize%
		Our synthetic truth modeled and rendered
		using Blender for the present work. The bottom images are sample frames
		of three different videos for different illumination conditions. A fourth
		sequence is also used in the experiments, mixing up frames from the three
		conditions.
	}
	\label{fig:synth:gt:pavilion:frames} \ReduceAfterCaptionfigspace
\end{figure}

Our first approach for reliable and fair evaluation of our 3D drawing algorithm
is to utilize a synthetic 3D model and a rendering software to factor out
calibration and reconstruction errors common among ground truth models obtained
from real world objects. Here, the realistically-rendered images for this scene,
Figure~\ref{fig:synth:gt:pavilion:frames}, as well as the precisely calibrated
views, are obtained using Blender. Three different illumination conditions were
rendered, and these can be mixed up to test any given algorithm's robustness
under varying illumination, such as a slow sunset. This synthetic data was
modeled after a real scene in Barcelona.

\begin{figure}
	\vspace{-3cm}
	\centering
	\includegraphics[width=1\linewidth]{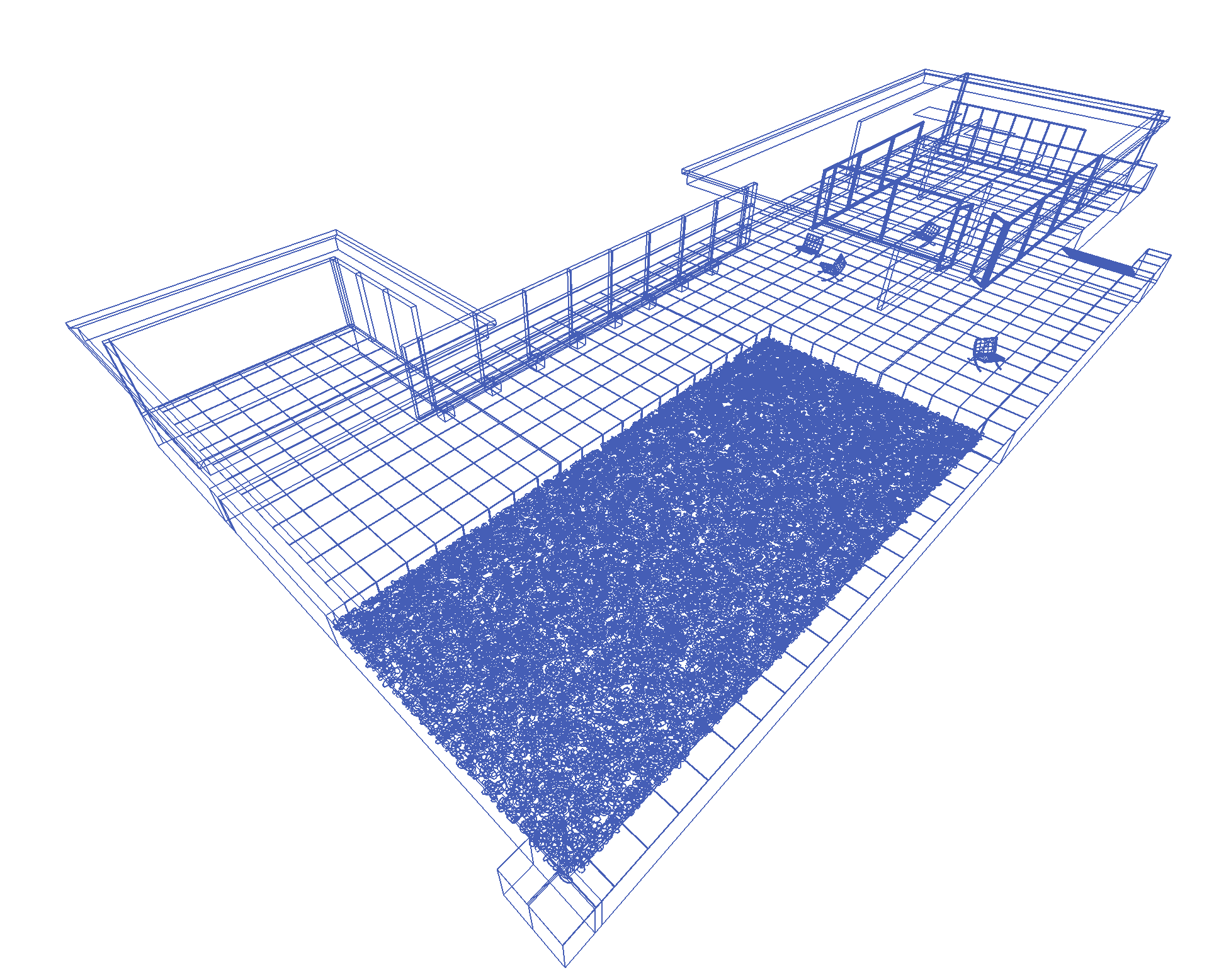}
	\includegraphics[width=1\linewidth]{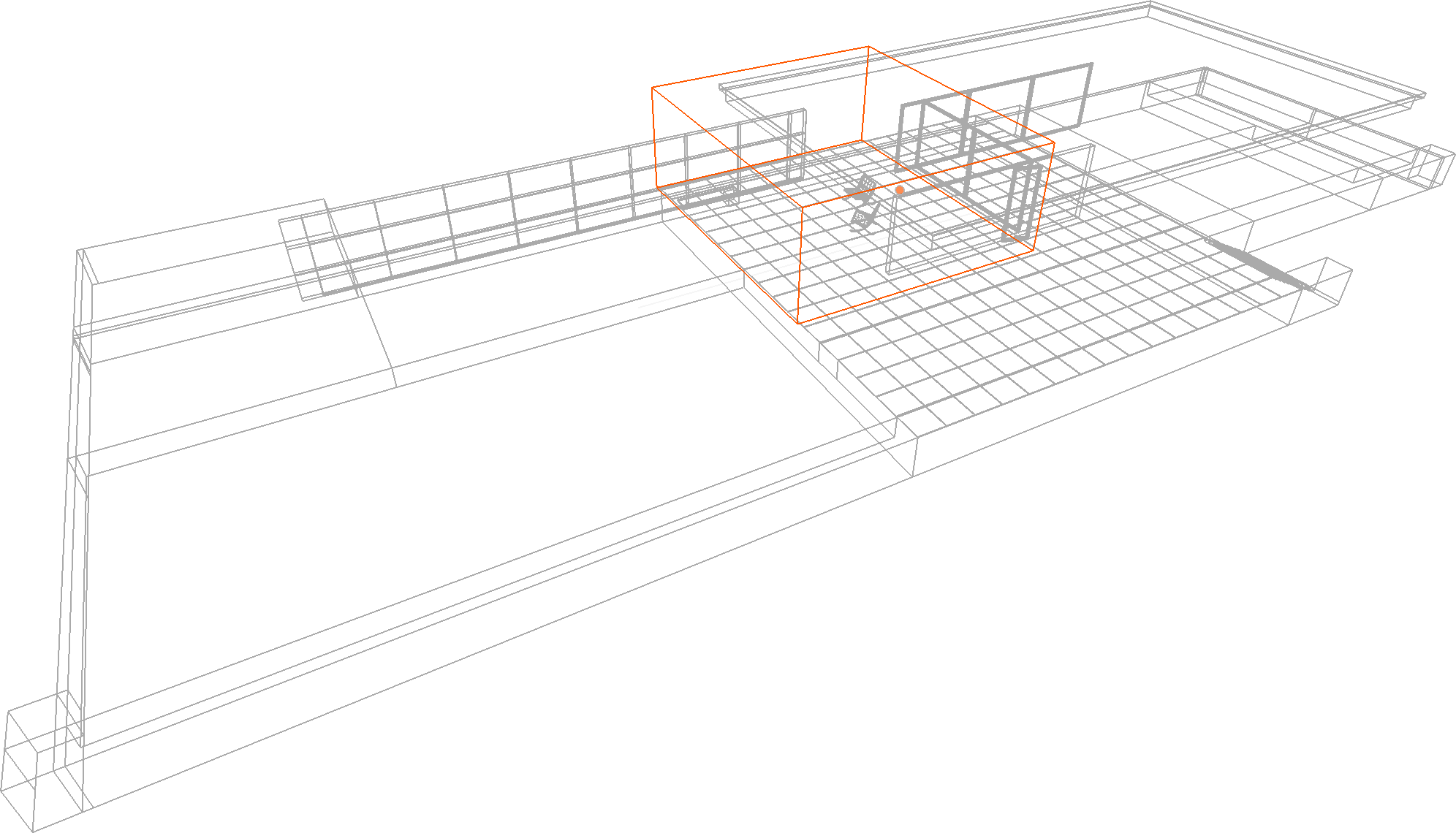}
	\ReduceBeforeCaptionfigspace
	\caption{\small %
		The full Barcelona Pavilion synthetic ground truth (top) and the bounding box
		(bottom) corresponding to Figure 10 in the paper.
	}
	\label{fig:synth:gt:pavilion:curves:fake} \ReduceAfterCaptionfigspace
\end{figure}

\begin{figure}
	\vspace{-2cm}
	\centering
	\includegraphics[width=\linewidth]{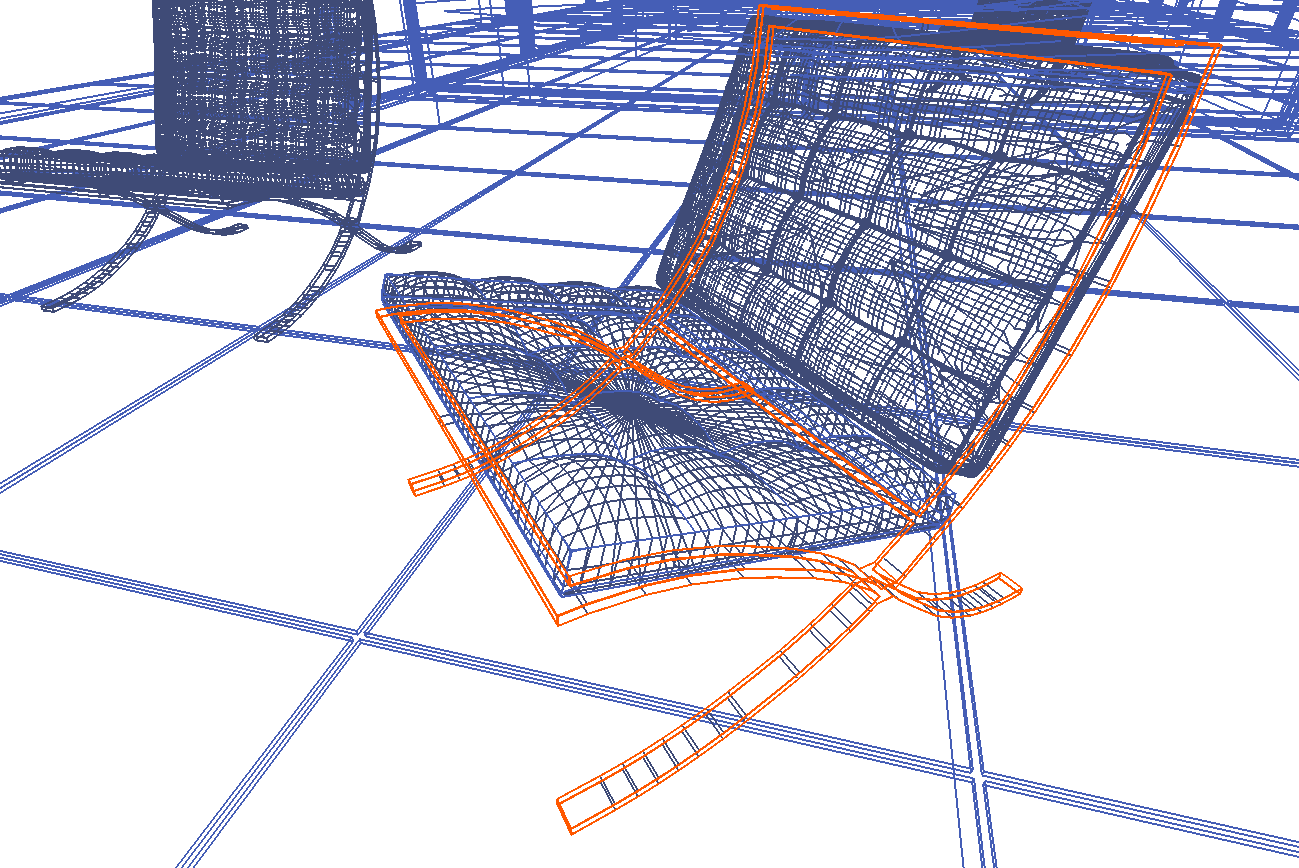}
	\includegraphics[width=\linewidth]{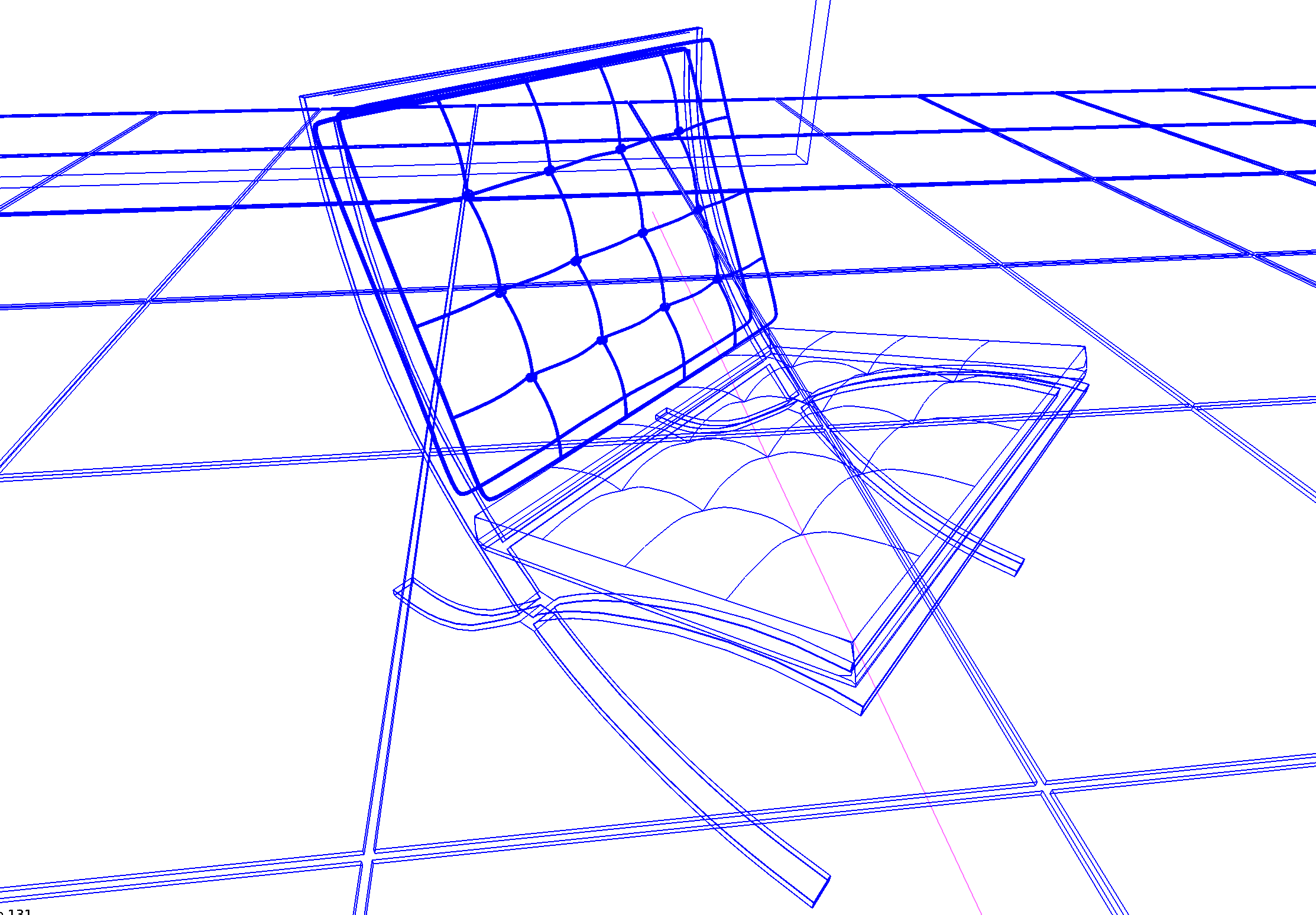}
	\ReduceBeforeCaptionfigspace
	\vspace{-0.7cm}
	\caption{\small %
		Process of deleting mesh edges to produce the desired ground truth edges.
	}
	\label{fig:synth:gt:pavilion:chair:overlay} \ReduceAfterCaptionfigspace
\end{figure}

\begin{figure}[h!]
	\centering
	\includegraphics[width=0.49\linewidth]{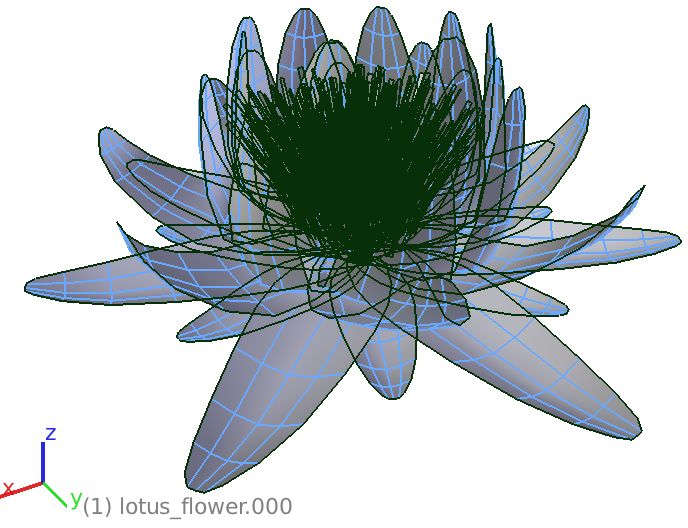}
	\includegraphics[width=0.49\linewidth]{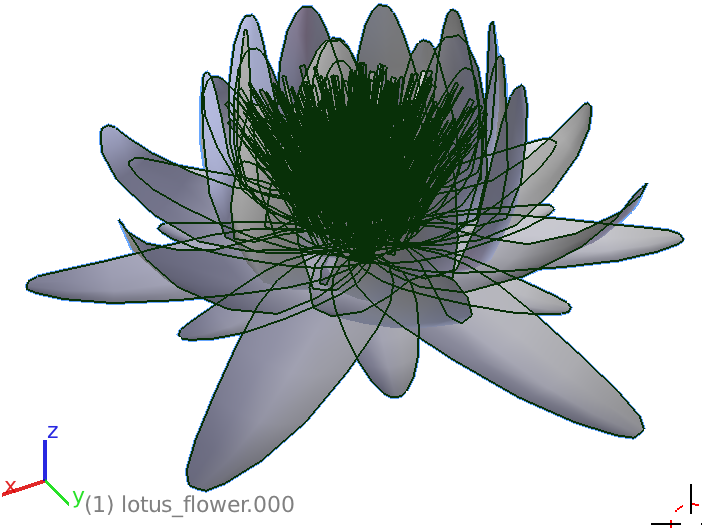}
	\includegraphics[width=0.49\linewidth]{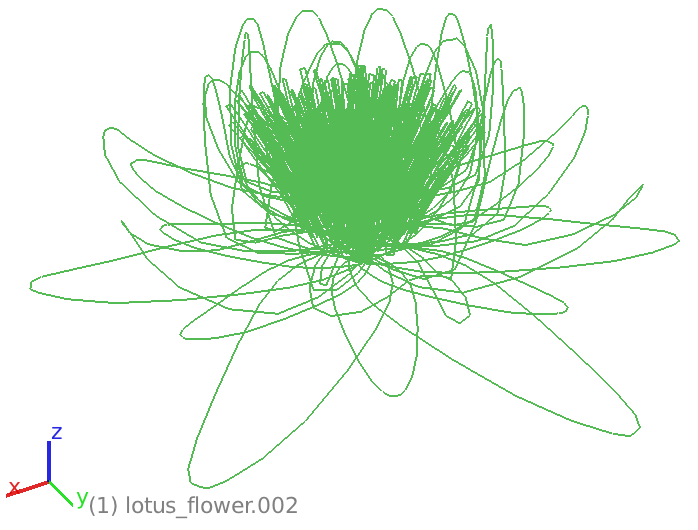}
	\ReduceBeforeCaptionfigspace
	\caption{\small %
		Detail of our ground truth generation. Even minute objects were modeled by
		discarding internal mesh edges (blue).
	}
	\label{fig:synth:gt:pavilion:lotus} \ReduceAfterCaptionfigspace
\end{figure}

To the best of our knowledge, there is no popular, publicly-available multiview
stereo ground truth that is based on a precise and complex 3D model and its
rendered images. We have made two versions of our Barcelona Pavilion dataset available for the evaluation of 3D reconstruction algorithms: i) The full mesh version for evaluating dense surface reconstruction algorithms, ii) 3D curve version for evaluating curvilinear models, such as the 3D drawing presented in this work, Figure~\ref{fig:synth:gt:pavilion:curves:fake}. The latter version was obtained by a Blender-aided process of manually deleting surface meshes until only the outline of the objects remained, see Figure~\ref{fig:synth:gt:pavilion:chair:overlay} and Figure~\ref{fig:synth:gt:pavilion:lotus}. 

Although the Barcelona Pavilion dataset allows for a very precise and reliable
way of evaluating 3D models, a point can be made about the necessity of testing
any reconstruction algorithm in the context of real world objects and real
camera imagery to get a real sense of its performance. Our second approach,
therefore, is to appropriate one of the many scenes present in DTU Robot
Dataset~\cite{Aan:Pedersen:etal:IJCV2012} to the task of evaluating 3D
curvilinear reconstructions. This is a significantly harder task than
eliminating the surface meshes in the synthetic case, since the ground truth
representation is a 3D point cloud, and no explicit distinction is made between
curve outlines and surface geometry. We therefore use Blender to project the 3D
point cloud ground truth for our selected scene onto several different images,
correct for calibration errors to the best of our capacity, then remove all the
internal surface points to end up with a subset of 3D points which are in the
proximity of curved structures in the scene, \textbf{see the full supplementary
materials}.

\afterpage{\FloatBarrier}
\section{Additional Results}\label{sec:moreresults}

\begin{figure}
	\centering
	\includegraphics[width=\linewidth]{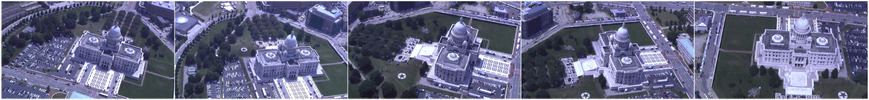}
	\includegraphics[width=1.1\linewidth]{figs/capitol-drawing-1.png}
	\includegraphics[width=1.1\linewidth]{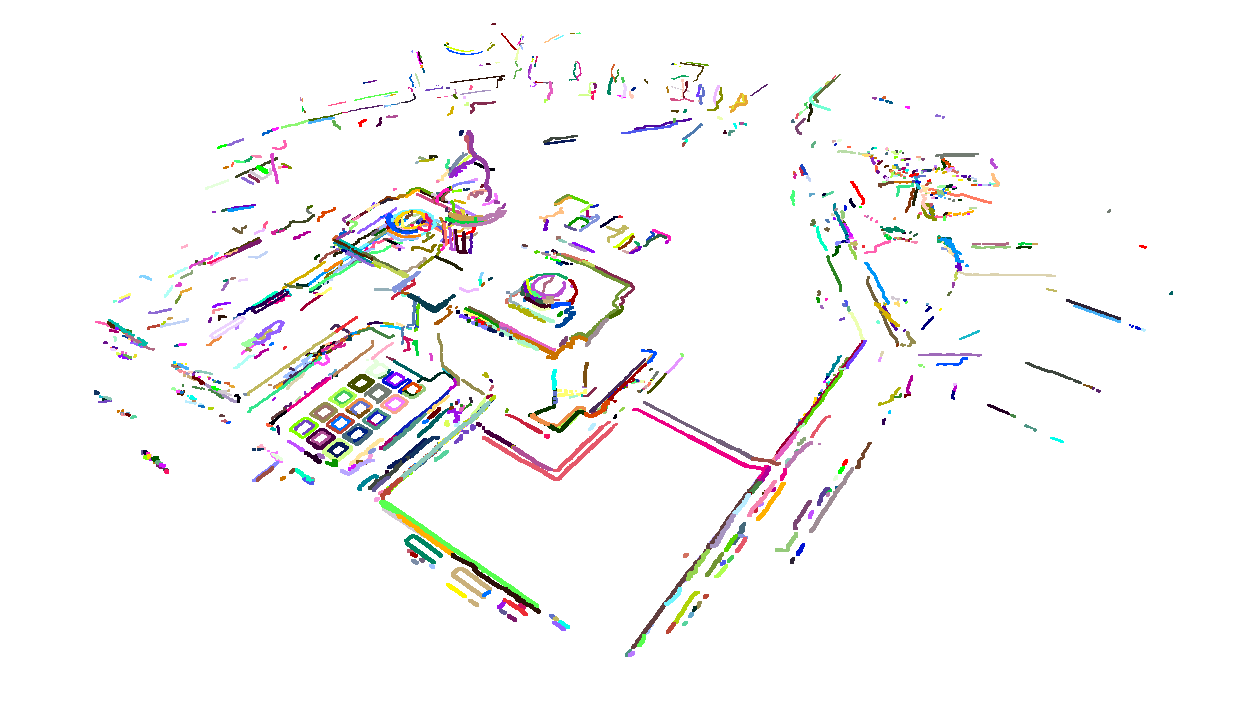}
	\ReduceBeforeCaptionfigspace
	\caption{\small %
		Curve drawing results for the Capitol High dataset.
	}
	\label{fig:cap:draw} 
	\ReduceAfterCaptionfigspace
\end{figure}

\begin{figure}
	\centering
	\includegraphics[width=\linewidth]{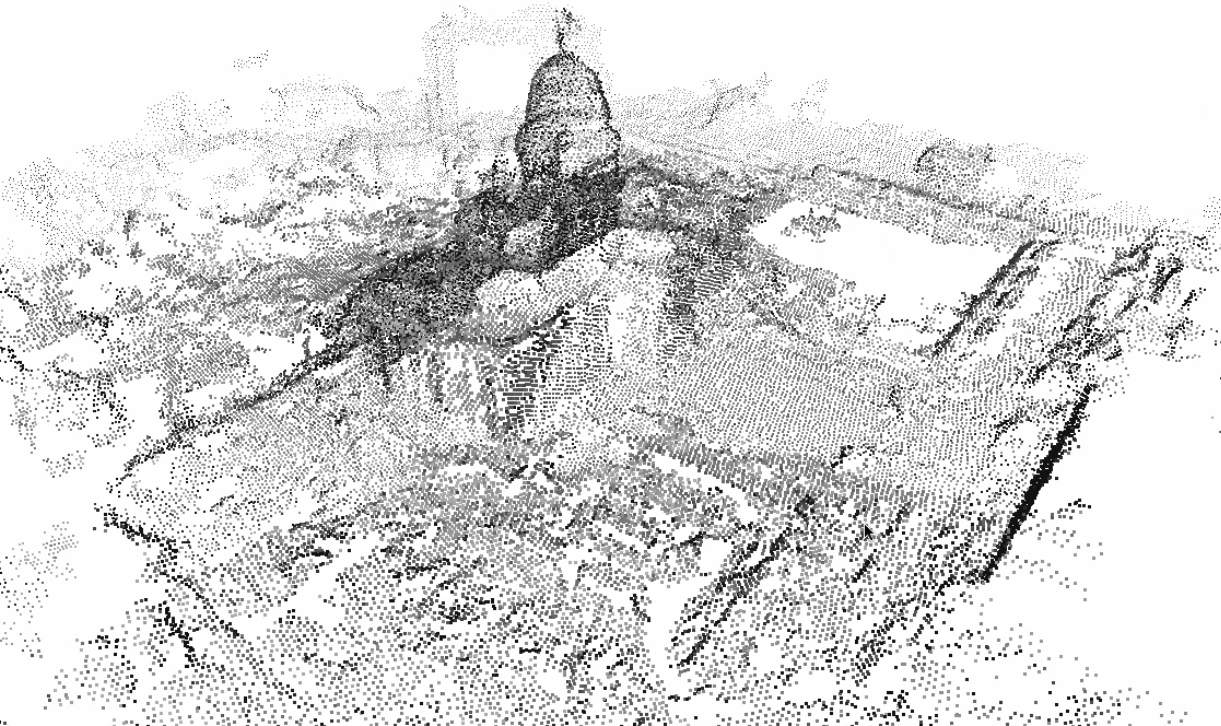}
	\includegraphics[width=\linewidth]{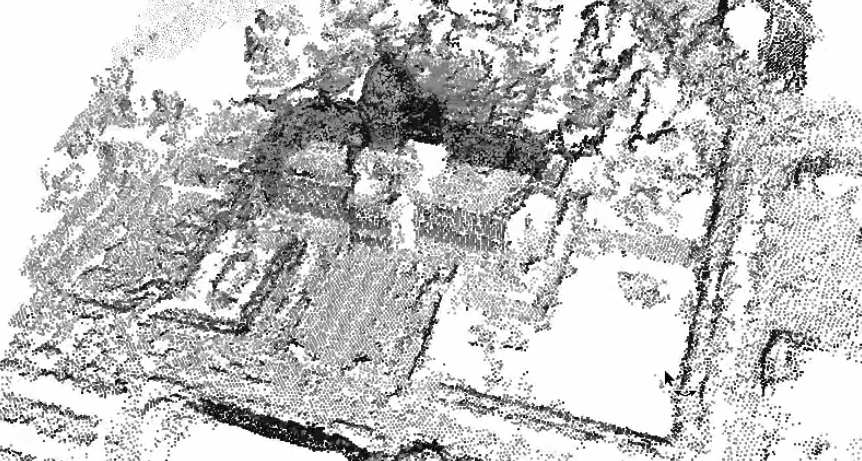}
	\ReduceBeforeCaptionfigspace
	\caption{\small %
		Reference PMVS results for the Capitol High dataset.
	}
	\label{fig:cap:pmvs} 
	\ReduceAfterCaptionfigspace
\end{figure}

In this Section we present more detailed figures for our results presented in
the main paper, Figure~\ref{fig:cap:draw}; as well as visual comparisons to the
results of PMVS~\cite{Furukawa:Ponce:CVPR2007}, Figure~\ref{fig:cap:pmvs}.
\textbf{See the full supplementary materials for additional results.}

\section{Additional Details}
\label{sec:details}
\subsection{Language} We used C++ to implement the base system up to the
enhanced curve sketch, using widely-available open source
libraries, such as Boost (\url{www.boost.org}), and VXL
(\url{vxl.sourceforge.net}). The curve drawing stage is implemented in Matlab.
The experiments ran on Linux but the code is very portable. 

\subsection{Additional Supplementary Material} Other than this pdf document,
\textbf{there is a full supplementary materials document}, as well as a
supplementary materials package which contains, among others: i) Two mp4 vieos comparing
reconstructions of Curve Sketch, Enhanced Curve Sketch, 3D Drawing and PMVS on
Amsterdam House Dataset, and ii) A .PLY file which contains the 3D
Drawing results on the Amsterdam House Dataset. You can view this model in
MeshLab or any other software that supports .PLY file format.

\subsection{Availability}
The C++ and Matlab source code are available at
\url{multiview-3d-drawing.sourceforge.net}, as well as the ground truth datasets
and additional supplementary material.

\small{
\paragraph{\textbf{Acknowledgements:}} 
This research received support from FAPERJ/Brazil E25/2014/204167,
UERJ/Brazil Prociencia 2014-2017, and NSF awards 1116140 and 1319914.
}

\bibliographystyle{splncs}
%\bibliography{egbib}
%\bibliographystyle{ieee}
%\bibliography{rf-multiview}

\begin{thebibliography}{10}\itemsep=-1pt

\bibitem{Abuhashim:Sukkarieh:IROS2012}
T.~Abuhashim and S.~Sukkarieh.
\newblock Incorporating geometric information into gaussian process terrain
  models from monocular images.
\newblock {\em IEEE/RSJ IROS'12}, pp. 4162--4168.

\bibitem{Baatz:Pollefeys:etal:ECCV12}
G.~Baatz, O.~Saurer, K.~K\"{o}ser, and M.~Pollefeys.
\newblock Large scale visual geo-localization of images in mountainous terrain.
\newblock ECCV'12.

\bibitem{Calakli:etal:3DIMPVT2012}
F.~Calakli, A.~O. Ulusoy, M.~I. Restrepo, G.~Taubin, and J.~L. Mundy.
\newblock High resolution surface reconstruction from multi-view aerial
  imagery.
\newblock {\em 3DIMPVT'12}, pp. 25--32.

\bibitem{Carlson:etal:ICRA2014}
F.~B. Carlson, N.~D. Vuong, and R.~Johansson.
\newblock Polynomial reconstruction of {3D} sampled curves using auxiliary
  surface data.
\newblock {\em IEEE ICRA'14}, pp. 4411--4416. 

\bibitem{Chen:Klette:IVT2014}
T.-Y. Chen and R.~Klette.
\newblock Animated non-photorealistic rendering in multiple styles.
\newblock {\em PSIVT'13}, p.\ 12--23. 2014.

\bibitem{Diskin:Vijayan:JEI2015}
Y.~Diskin and V.~Asari.
\newblock Dense point-cloud representation of a scene using monocular vision.
\newblock {\em Journal of Electronic Imaging}, 24(2), 2015.

%\bibitem{Diskin:Asari:SPIEDSS2012}
%Y.~Diskin and V.~K. Asari.
%\newblock Dense point-cloud creation using superresolution for a monocular {3D}
%  reconstruction system.
%\newblock {\em Defense, Security, and Sensing}, SPIE, 2012.

%\bibitem{Diskin:Asari:ICSMC2013}
%Y.~Diskin and V.~K. Asari.
%\newblock {3D} scene reconstruction for aiding unmanned vehicle navigation.
%\newblock {\em IEEE Conference on Systems, Man, and Cybernetics}, pp. 243--248.2013.

\bibitem{Fabbri:PhD:2010}
R.~Fabbri.
\newblock {\em Multiview Differential Geometry in Application to Computer
  Vision}.
\newblock {Ph.D.} dissertation, Division Of Engineering, Brown University, 2010.

\bibitem{Fabbri:Giblin:Kimia:ECCV12}
R.~Fabbri, P.~J. Giblin, and B.~B. Kimia.
\newblock Camera pose estimation using first-order curve differential geometry.
\newblock {\em ECCV'12}.

\bibitem{Fabbri:Kimia:EMMCVPR2005}
R.~Fabbri and B.~B. Kimia.
\newblock High-order differential geometry of curves for multiview
  reconstruction and matching.
\newblock {\em EMMCVPR'05}, pp. 645--660.

\bibitem{Fabbri:Kimia:CVPR10}
R.~Fabbri and B.~B. Kimia.
\newblock {3D} curve sketch: Flexible curve-based stereo reconstruction and
  calibration.
\newblock {\em CVPR}, 2010.

\bibitem{Fathi:etal:AEI2015}
H.~Fathi, F.~Dai, and M.~Lourakis.
\newblock Automated as-built {3D} reconstruction of civil infrastructure using
  computer vision.
\newblock {\em Advanced Engineering Informatics}, 2015.

\bibitem{Feng:Medioni:etal:SIGGRAPH14}
A.~Feng, A.~Shapiro, W.~Ruizhe, M.~Bolas, G.~Medioni, and E.~Suma.
\newblock Rapid avatar capture and simulation using commodity depth sensors.
\newblock SIGGRAPH'14, 2014.

\bibitem{Fleming:etal:PNAS2011}
R.~W. Fleming, D.~Holtmann-Rice, and H.~H. B{\"u}lthoff.
\newblock Estimation of {3D} shape from image orientations.
\newblock {\em PNAS}, 108(51):20438--20443, 2011.

\bibitem{Yuliang:etal:CVPR14}
Y.~Guo, N.~Kumar, M.~Narayanan, and B.~Kimia.
\newblock A multi-stage approach to curve extraction.
\newblock {\em CVPR'14}.

\bibitem{Heinly:Frahm:etal:CVPR2015}
J.~Heinly, J.~L. Sch\"{o}nberger, E.~Dunn, and J.-M. Frahm.
\newblock Reconstructing the world in six days.
\newblock {\em CVPR'15}.

\bibitem{Jensen:etal:CVPR14}
R.~Jensen, A.~Dahl, G.~Vogiatzis, E.~Tola, and A.~s~H.
\newblock Large scale multi-view stereopsis evaluation.
\newblock {\em CVPR'14}.

\bibitem{Koenderink:Wagemans:etal:iPerception2013}
J.~Koenderink, A.~van Doorn, and J.~Wagemans.
\newblock {SfS?} not likely\dots.
\newblock {\em i-Perception}, 4(5):299, 2013.

\bibitem{Kowdle:etal:ECCV10}
A.~Kowdle, D.~Batra, W.-C. Chen, and T.~Chen.
\newblock imodel: Interactive co-segmentation for object of interest {3D}
  modeling.
\newblock ECCV'12, pp. 211--224.

\bibitem{Kunsberg:Zucker:LNM2014}
B.~Kunsberg and S.~W. Zucker.
\newblock Why shading matters along contours.
\newblock {\em Neuromathematics of Vision}, pp. 107--129. 2014.

\bibitem{Lebeda:etal:ACCV2014}
K.~Lebeda, S.~Hadfield, and R.~Bowden.
\newblock {2D} or not {2D}: Bridging the gap between tracking and structure
  from motion.
\newblock {\em ACCV'14}.

\bibitem{Litvinov:etal:IC3D2012}
V.~Litvinov, S.~Yu, and M.~Lhuillier.
\newblock 2-manifold reconstruction from sparse visual features.
\newblock {\em IEEE IC3D'12}, pp.\ 1--8.

\bibitem{Mattingly:etal:JVLC2015}
W.~A. Mattingly, J.~H. Chariker, R.~Paris, D.~jen Chang, and J.~R. Pani.
\newblock {3D} modeling of branching structures for anatomical instruction.
\newblock {\em Journal of Visual Languages \& Computing}, 29(0):54 -- 62, 2015.

\bibitem{Wendel:etal:CVWW2011}
K.~P\"{o}tsch and A.~Pinz.
\newblock {3D} geometric shape modeling by `{3D} contour cloud’
  reconstruction from stereo videos.
\newblock {\em Computer Vision Winter Workshop}, p.~99. 2011.

\bibitem{Rao:etal:IROS2012}
D.~Rao, S.-J. Chung, and S.~Hutchinson.
\newblock {CurveSLAM}: An approach for vision-based navigation without point
  features.
\newblock In {\em IEEE/RSJ IROS}, pp.\ 4198--4204, 2012.

\bibitem{Restrepo:etal:JPRS2014}
M.~Restrepo, A.~Ulusoy, and J.~Mundy.
\newblock Evaluation of feature-based 3-{D} registration of probabilistic
  volumetric scenes.
\newblock {\em Journal of Photogrammetry and Remote Sensing}, 98:1--18,
  2014.

\bibitem{Carrasco:etal:LNCS2012}
J.~R.~Espino, J.-J. Gonzalez-Barbosa, R.~G.~Loenzo, D.~C.~Esparza,
  and R.~Gonzalez-Barbosa.
\newblock Vision system for {3D} reconstruction with telecentric lens.
\newblock  {\em LNCS} 7329, 2012.

\bibitem{Shinozuka:Saito:VRIC14}
Y.~Shinozuka and H.~Saito.
\newblock Sharing {3D} object with multiple clients via networks using
  vision-based {3D} object tracking.
\newblock VRIC'14, pp. 34:1--34:4

\bibitem{Simoes:etal:SVR2014}
F.~Simoes, M.~Almeida, M.~Pinheiro, and R.~dos Anjos.
\newblock Challenges in {3D} reconstruction from images for difficult
  large-scale objects.
\newblock {\em SVAR'14}, 0:74--83.

\bibitem{Teney:Piater:3DIMPVT12}
D.~Teney and J.~Piater.
\newblock Sampling-based multiview reconstruction without correspondences for
  {3D} edges.
\newblock {\em 3DIMPVT'12}, pp. 160--167.

\bibitem{Ruizhe:Medioni:CVPR2014}
R.~Wang, J.~Choi, and G.~Medioni.
\newblock {3D} modeling from wide baseline range scans using contour coherence.
\newblock {\em CVPR'14}.

\bibitem{Zhang:line:PHDThesis2013}
L.~Zhang.
\newblock {\em Line Primitives and Their Applications in Geometric Computer
  Vision}.
\newblock PhD thesis, 2013.

\bibitem{Zia:Stark:Schindler:IJCV2015}
M.~Zia, M.~Stark, and K.~Schindler.
\newblock Towards scene understanding with detailed {3D} object
  representations.
\newblock {\em IJCV}, 112(2):188--203,
  2015.

\bibitem{Zucker:PIEEE2014}
S.~Zucker.
\newblock Stereo, shading, and surfaces: Curvature constraints couple neural
  computations.
\newblock {\em Proc.\ IEEE}, 102(5):812--829, 2014.

\bibitem{Argarwal:Snavely:etal:ICCV09}
S.~Agarwal, N.~Snavely, I.~Simon, S.~M. Seitz, and R.~Szeliski.
\newblock Building {Rome} in a day.
\newblock {\em ICCV'09}.

%@article{aanæsinteresting,
%  title={Interesting Interest Points},
%  author={Aan{\ae}s, H. and Dahl, A.L. and Steenstrup Pedersen, K.},
%  journal={International Journal of Computer Vision},
%  pages={18--35},
%  year={2012},
%  volume={97},
%  publisher={Springer}
%}
\bibitem{Aan:Pedersen:etal:IJCV2012}
H.~Aan{\ae}s, A.L.~Dahl, and K.~Pedersen
\newblock Interesting Interest Points
\newblock {\em IJCV}, 2012.

%\bibitem{Astrom:Cipolla:Giblin:IJCV1999}
%K.~Astrom, R.~Cipolla, and P.~Giblin.
%\newblock Generalised epipolar constraints.
%\newblock {\em IJCV}, 33(1):51--72, 1999.

\bibitem{Berthilsson:etal:IJCV2001}
R.~Berthilsson, K.~{\AA}str{\"o}m, and A.~Heyden.
\newblock Reconstruction of general curves, using factorization and bundle
  adjustment.
\newblock {\em IJCV}, 41(3):171--182, 2001.

%\bibitem{Bottino:Laurentini:PAMI2004}
%A.~Bottino and A.~Laurentini.
%\newblock The visual hull of smooth curved objects.
%\newblock {\em PAMI},
%  26(12):1622--1632, 2004.

\bibitem{Giblin:Motion:Book}
R.~Cipolla and P.~Giblin.
\newblock {\em Visual Motion of Curves and Surfaces}.
\newblock Cambridge, 1999.

\bibitem{Cole:etal:SIGGRAPH09}
F.~Cole, K.~Sanik, D.~DeCarlo, A.~Finkelstein, T.~Funkhouser, S.~Rusinkiewicz,
  and M.~Singh.
\newblock How well do line drawings depict shape?
\newblock {\em SIGGRAPH'09}.

\bibitem{Crispell:etal:LNCS2009}
D.~Crispell, D.~Lanman, P.~Sibley, Y.~Zhao, and G.~Taubin.
\newblock Shape from depth discontinuities.
\newblock {\em LNCS} 5416, 2009.

%\bibitem{Ma:IJCV93}
%S.~De~Ma.
%\newblock Conics-based stereo, motion estimation, and pose determination.
%\newblock {\em IJCV}, 10(1):7--25, 1993.

\bibitem{Furukawa:Ponce:CVPR2007}
Y.~Furukawa and J.~Ponce.
\newblock Accurate, dense, and robust multi-view stereopsis.
\newblock {\em CVPR'07}.

\bibitem{Goesele:etal:ICCV07}
M.~Goesele, N.~Snavely, B.~Curless, H.~Hoppe, and S.~Seitz.
\newblock Multi-view stereo for community photo collections.
\newblock {\em ICCV'07}.

\bibitem{Habbecke:Kobbelt:CVPR2007}
M.~Habbecke and L.~Kobbelt.
\newblock A surface-growing approach to multi-view stereo reconstruction.
\newblock {\em CVPR'07}.

%\bibitem{Hartley:Zisserman:multiple:view}
%R.~Hartley and A.~Zisserman.
%\newblock {\em Multiple View Geometry in Computer Vision}.
%\newblock Cambridge, 2000.

%\bibitem{Hernandez:etal:PAMI07}
%C.~Hernandez, F.~Schmitt, and R.~Cipolla.
%\newblock Silhouette coherence for camera calibration under circular motion.
%\newblock {\em PAMI}, 29(2):343--349. 2007

\bibitem{Hernandez:Schmitt:CVIU04}
C.~H. Esteban and F.~Schmitt.
\newblock Silhouette and stereo fusion for 3{D} object modeling.
\newblock {\em CVIU}, 96(3):367--392. 2004

\bibitem{Leggitt:drawing:book}
J.~Leggitt
\newblock {\em Drawing Shortcuts: Developing Quick Drawing Skills Using Today's Technology}.
\newblock Wiley, 2015.

%\bibitem{Jain:etal:Tracking:CVIU07}
%V.~Jain, B.~B. Kimia, and J.~L. Mundy.
%\newblock Segregation of moving objects using elastic matching.
%\newblock {\em CVIU}, 108:230--242. 2007

%\bibitem{Kahl:Heyden:ICCV98}
%F.~Kahl and A.~Heyden.
%\newblock Using conic correspondence in two images to estimate the epipolar
%  geometry.
%\newblock {\em ICCV'98}, p.\ 761.

%\bibitem{Kaminski:Shashua:2004}
%J.~Y. Kaminski and A.~Shashua.
%\newblock Multiple view geometry of general algebraic curves.
%\newblock {\em IJCV}, 56(3):195--219. 2004

\bibitem{Liu:Cooper:etal:PAMI07}
S.~Liu, K.~Kang, J.-P. Tarel, and D.~B. Cooper.
\newblock Freeform object reconstruction from silhouettes, occluding edges and
  texture edges
\newblock {\em PAMI}, 2007.

\bibitem{Moreels:Perona:IJCV07}
P.~Moreels and P.~Perona.
\newblock Evaluation of features detectors and descriptors based on 3d objects.
\newblock {\em IJCV}, 73(3):263--284, 2007.

\bibitem{Pollefeys:VanGool:etal:handheld:IJCV2004}
M.~Pollefeys, L.~V. Gool, M.~Vergauwen, F.~Verbiest, K.~Cornelis, J.~Tops, and
  R.~Koch.
\newblock Visual modeling with a hand-held camera.
\newblock {\em IJCV}, 59(3):207--232, 2004.

%\bibitem{Porrill:Pollard:1991}
%J.~Porrill and S.~Pollard.
%\newblock Curve matching and stereo calibration.
%\newblock {\em IVC}, 9(1):45--50, 1991.

%\bibitem{Reyes:Corrochano:IVC05}
%L.~Reyes and E.~Bayro~Corrochano.
%\newblock The projective reconstruction of points, lines, quadrics, plane
%  conics and degenerate quadrics using uncalibrated cameras.
%\newblock {\em IVC}, 23(8):693--706. 2005.

%\bibitem{Robert:Faugeras:1991}
%L.~Robert and O.~D. Faugeras.
%\newblock Curve-based stereo: figural continuity and curvature.
%\newblock {\em CVPR'91}.

%\bibitem{Schmid:Zisserman:2000}
%C.~Schmid and A.~Zisserman.
%\newblock The geometry and matching of lines and curves over multiple views.
%\newblock {\em IJCV}, 40(3):199--233, 2000.

\bibitem{Seitz:etal:CVPR06}
S.~Seitz, B.~Curless, J.~Diebel, D.~Scharstein, and R.~Szeliski.
\newblock A comparison and evaluation of multi-view stereo reconstruction
  algorithms.
\newblock {\em CVPR'06}.

\bibitem{Strecha:etal:CVPR08}
C.~Strecha, W.~von Hansen, L.~Van Gool, P.~Fua, and U.~Thoennessen 
\newblock On Benchmarking Camera Calibration and Multi-View Stereo for High Resolution Imagery
\newblock {\em CVPR'08}


\bibitem{Tamrakar:Kimia:ICCV07}
A.~Tamrakar and B.~B. Kimia.
\newblock No grouping left behind: From edges to curve fragments.
\newblock {\em ICCV '07}.

\bibitem{Yee:architectural:book}
R.~Yee
\newblock {\em Architectural Drawing: A Visual Compendium of Types and Methods}.
\newblock Wiley, 2012.

\end{thebibliography}

\end{document}